\documentclass[lettersize, journal]{IEEEtran}
\usepackage{amsmath,amsfonts}
\usepackage{array}
\usepackage[caption=false,font=normalsize,labelfont=sf,textfont=sf]{subfig}
\usepackage{textcomp}
\usepackage{stfloats}
\usepackage{url}
\usepackage{verbatim}
\usepackage{graphicx}
\usepackage{cite}
\usepackage{xcolor}
\usepackage{algorithmicx}
\usepackage{algorithm,algpseudocode}
\usepackage{amsfonts,amssymb}
\usepackage{bbding}
\usepackage{bm}
\usepackage{booktabs}
\usepackage{multirow}

\usepackage{tabularx}

\usepackage{gensymb}

\begin{document}
\title{Real-time Monocular 2D and 3D Perception of Endoluminal Scenes for Controlling Flexible \\ Robotic Endoscopic Instruments}

\author{Ruofeng~Wei,~\IEEEmembership{Member, IEEE},~Kai~Chen,~\IEEEmembership{Member, IEEE},~Yui-Lun~Ng,~Yiyao~Ma,~\IEEEmembership{Student Member, IEEE},\\~Justin~Di-Lang~Ho,~Hon-Sing~Tong,~Xiaomei~Wang,~\IEEEmembership{Member, IEEE},~Jing~Dai,~\IEEEmembership{Member, IEEE},\\~Ka-Wai~Kwok$\textsuperscript{\textdagger}$,~\IEEEmembership{Senior Member, IEEE},~and~Qi~Dou$\textsuperscript{\textdagger}$,~\IEEEmembership{Senior Member, IEEE}
\thanks{This work was supported in part by Hong Kong Innovation and Technology Commission under Project No. PRP/026/22FX, in part by a grant from the NSFC/RGC Joint Research Scheme sponsored by the Research Grants Council of the Hong Kong Special Administrative Region, China and the National Natural Science Foundation of China (Project No. N\_CUHK410/23), in part by the HK RGC AoE under AoE/E-407/24-N, in part by the National Natural Science Foundation of China Project No.~62322318, and in part by Agilis Robotics and its subsidiaries Agilis Robotics Limited and Agilis Robotics Limited (Guangzhou).
The authors are deeply thankful to Mr. Joe K.M. Hui and Mr. King S.M. Wong for positioning the presented technology in the appropriate applications and markets.
(\emph{Corresponding authors: Ka-Wai Kwok and Qi Dou})}%
\thanks{Ruofeng Wei, Kai Chen, Yiyao Ma, and Qi Dou are with the Department of Computer Science and Engineering, The Chinese University of Hong Kong, Hong Kong, China.}
\thanks{Yui-Lun Ng, Justin Di-Lang Ho, Hon-Sing Tong, Xiaomei Wang, and Jing Dai are with the Department of Mechanical
Engineering, The University of Hong Kong, China.}
\thanks{Ka-Wai Kwok is with the Department of Mechanical and Automation
Engineering, The Chinese University
of Hong Kong, and also with the Department of Mechanical
Engineering, The University of Hong Kong.}
}



\maketitle

\begin{abstract}
Endoluminal surgery offers a minimally invasive option for early-stage gastrointestinal and urinary tract, but is limited by basic surgical tools and a steep learning curve. Robotic systems, particularly continuum robots, provide flexible instruments that enable precise, intuitive tissue resection in confined spaces, potentially improving outcomes. This paper presents an integrated visual perception platform for a continuum robotic system in endoluminal surgery. Our objective is to leverage monocular endoscopic image-based perception algorithms to accurately identify the position and orientation of flexible instruments and measure their distances from surrounding tissues. This thorough understanding of continuum robots and surgical scenes enhances the robustness of robotic procedures. We introduce 2D and 3D learning-based perception algorithms and develop a physically-realistic simulator that models the dynamics of flexible instruments. This simulator features a pipeline for generating realistic endoluminal scenes, enabling control of flexible robots in a realistic environment and substantial data collection. Using a continuum robot prototype, we conducted extensive evaluations, including module assessments and system-level evaluation of the perception platform. Results demonstrate that our perception algorithms significantly improve control of flexible instruments, reducing manipulation time by over 70\% for trajectory-following tasks and enhancing the understanding of complex surgical scenarios, leading to robust endoluminal surgeries.
\end{abstract}


\section{Introduction}\label{sec:introduction}

\subsection{Background and Motivation}

\IEEEPARstart{E}{ndoluminal} surgery through natural orifices (e.g. mouth, anus, urethra) offers an effective, incisionless treatment for early-stage cancers in the gastrointestinal (GI) and urinary tracts.
However, current practice suffers from basic surgical instruments that lack dexterity and effective tissue retraction capabilities, resulting in poor tumor resection quality~\cite{oude2018intuitive,ohuchida2020robotic}. Patient outcomes depend heavily on clinicians' experience due to the significant technical skill required to precisely maneuver endoscopic instruments, particularly long and flexible GI endoscopes~\cite{lee2021easyendo}. These factors create a steep learning curve and poor resection rates, ultimately leading to high tumor recurrence~\cite{zhang2020learning}.

\begin{figure}[tp]
\centering
\includegraphics[width = 1.0\hsize]{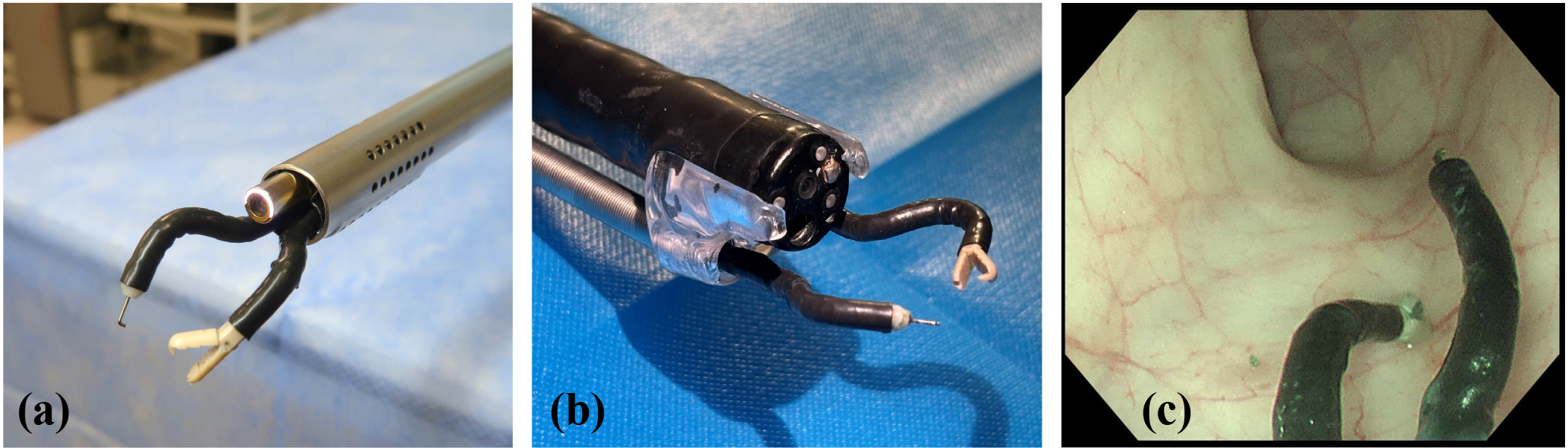}
\caption{Robotic instrument configurations and endoscopic scenarios in continuum robotic endoluminal surgery. (a) For transurethral bladder tumour resection, the robotic instruments are deployed through a standard urology outer sheath alongside a standard telescope.  (b) Two external instrument channels equipped to a standard single-channel GI endoscope (e.g. from Olympus) to deliver the robotic instruments. (c) Monocular endoluminal image captured by GI endoscope.}
\label{fig:agilis_robot}
\vspace{-0.3cm}
\end{figure}

Robotic systems have recently emerged to address these challenges by providing dexterous and flexible instruments that enable effective bimanual tissue resection. 
These systems allow surgeons to intuitively resect tissue with triangulation and tissue retraction capabilities previously limited to conventional open or laparoscopic surgeries~\cite{gao2024transendoscopic}.
Continuum robots represent a promising solution for endoluminal robotic instruments. Unlike conventional robots consisting of finite rigid links and joints~\cite{greer2019soft,ferguson2024unified} that are difficult to miniaturize and lack inherent flexibility, continuum robots feature curvilinear structures with infinite elastic joints, enabling continuous deformation and high flexibility—particularly advantageous in confined spaces~\cite{da2020challenges,xu2024shape}. Their dexterity and adaptability make continuum robots favorable for less traumatic endoluminal surgery~\cite{wu2024review,mao2024magnetic}, facilitating delicate operations on anatomical structures through natural tortuous orifices.
A common configuration includes two flexible instruments paired with a monocular endoscope~\cite{gao2024transendoscopic}, as shown in Fig.~\ref{fig:agilis_robot}(a) and (b).

Despite their flexibility, integrating continuum robots as surgical instruments presents numerous challenges. The lack of rigid coupling between actuators and end effectors, combined with redundant design, complicates kinematic modeling. Additionally, their deformable structures hinder three-dimensional (3D) intraoperative real-time shape perception, making accurate robot state modeling difficult~\cite{gao2024body}. Consequently, achieving precise and reliable motion control during surgical procedures becomes challenging. When surgeons manipulate flexible instruments in constrained endoluminal environments, collisions with surrounding anatomical structures are inevitable. The lack of geometric understanding between the robot and its surroundings further complicates control. Therefore, accurate perception of continuum robots and anatomical structures is essential for effective use of these systems in endoluminal surgery~\cite{wu2024review}.

Current continuum robot perception relies on intraoperative sensors such as Fiber Bragg Gratings (FBG) and Electromagnetic (EM) sensors~\cite{ha2022shape} integrated directly into the robot. However, these sensors are costly, and their functionality is often environment-constrained. Moreover, embedding sensors within the robot may reduce lifespan and introduce safety risks, making them unsuitable for critical scenarios like endoluminal surgery. Medical imaging systems such as fluoroscopy, ultrasound, Computed Tomography (CT)~\cite{goh2024intraoperative}, or Magnetic Resonance Imaging (MRI)~\cite{dai2023automatic}, which are external to the patient, are used to perceive anatomical structures, but their high cost restricts geometric measurements from multi-modal data~\cite{wu2024review}. Given these limitations, vision-based perception presents a promising alternative. Most flexible surgical robotic platforms primarily utilize monocular rather than stereo endoscopes to observe anatomies and tumors deep within the human body. This choice is driven by the confined operating space of surgical sites and size constraints of miniaturized robots. Thus, developing a monocular endoscopic image-based perception framework for continuum robotic systems is crucial for enhancing robotic endoluminal surgery.

\subsection{Challenges}

Current image-based perception frameworks for robotic surgery typically focus on single vision-based techniques, such as instrument segmentation~\cite{wang2024video} or 3D reconstruction of anatomical structure~\cite{wei2022stereo}. However, robotic endoluminal surgery presents fundamentally different requirements that demand a more comprehensive approach. When controlling flexible instruments to precisely and safely perform procedures like lesion removal along defined margins, surgeons must simultaneously perceive multiple critical elements: the two-dimensional (2D) position of instruments in endoscopic images, the 3D states of continuum robots, and the 3D geometric structures of the surrounding anatomical environment. This multi-modal perception is essential for minimizing harm to adjacent healthy tissue and ensuring procedural success. The challenge lies in developing an integrated framework that can accurately identify and track continuum robot positions and states while simultaneously capturing the complex 3D geometry of the entire endoscopic environment in real-time. 

The development of robust perception algorithms requires massive amounts of domain-specific data with corresponding labels~\cite{incetan2021vr}, which presents a significant challenge in the specialized field of robotic endoluminal surgery. While synthetic data generation has shown promise in improving learning-based vision methods and addressing the difficulty of obtaining real measurements~\cite{qiu2016unrealcv}, existing simulation platforms fall short of meeting the unique requirements of continuum robotics. Current available platforms such as VisionBlender~\cite{cartucho2021visionblender} and AMBF~\cite{munawar2019real} can generate labeled images of surgical tools in tissue backgrounds, but they primarily focus on robotic laparoscopy with rigid surgical instruments. The endoluminal surgical environment presents distinct characteristics: highly flexible continuum instruments, dynamic and narrow anatomical spaces, and complex tool-tissue interactions that differ significantly from laparoscopic procedures. This gap necessitates the development of realistic simulation environments specifically designed for flexible instruments and endoluminal scenarios to enable effective synthetic data generation for training robust perception algorithms.

\subsection{Contributions}
To our knowledge, no existing work has comprehensively addressed the perception challenges in continuum robotic systems using solely monocular vision. 
This paper presents an integrated 2D and 3D monocular perception framework for such systems. As illustrated in Fig.~\ref{fig:agilis_robot}(c), our method operates on standard endoscopic images that capture both instruments and dynamic surgical scenes simultaneously within a single frame. 
Specifically, a novel flexible robot segmentation method is developed, which resorts to robust feature representation from pre-trained vision foundation models and harnesses various surgical images in an annotation-efficient way to significantly improve the segmentation accuracy in complex surgical scenarios. Based on the 2D robotic instruments segmentation mask, we further employ a probabilistic model to represent the 3D states of the flexible instruments and develop an efficient network to jointly estimate robot state parameters along with their corresponding uncertainty from the endoscopic images. Additionally, we introduce a novel monocular depth foundation model to perceive the 3D geometry of both tissue surfaces and flexible robots.
Our method incorporates the illumination modeling and an expressive scene representation that fully considers light direction, illumination attenuation, and tissue's normal direction.
This approach effectively recovers relative depth to measure the instrument-tissue distance during surgery.
To generate large-scale endoscopic data for training our perception model, we construct a physically-realistic simulation environment that models the continuum robotic system. Finally, we integrate our image-based perception framework into a novel continuum robotic system to evaluate the accuracy and efficiency of our algorithms. In this work, the robotic instruments were delivered through two external instrument channels equipped to a standard single-channel GI endoscope by Olympus, as illustrated in Fig.~\ref{fig:agilis_robot}(b). The major contributions of this work are summarized as follows:
\begin{enumerate}
\item Development of an integrated monocular endoscopic image-based perception framework for endoluminal surgery, capable of providing real-time feedback to enhance control of surgical instruments for a continuum robotic system.
\item Design of a physically-realistic simulation platform to generate synthetic data, such that the simulated flexible instruments closely match the real instruments' configuration, and the surgical scene is highly realistic for the learning-based method training.
\item  Experimental evaluations on the perception framework quantitatively and qualitatively, including 2D segmentation, 3D robot state estimation, and 3D depth estimation modules. The perception framework was also validated in the continuum robotic system, allowing for flexible robot control with improved understanding of complex surgical scenarios, demonstrating its potential in robust endoluminal surgeries. 
\end{enumerate}
\section{Related Works} \label{sec:related_work}

\subsection{Robot State Estimation}
Most current approaches for estimating the state of continuum robots heavily rely on external sensors. 
These sensors collect data about the robot and its surrounding environment, which is then used to calculate the robot state parameters.
For example, some methods monitor changes in resistance or electric fields due to bending or force and pressure, and utilize them to compute robot state~\cite{resistive_sensor2,capacitive_sensor1}. 
Others leverage changes in the spectrum profile or magnetic flux to detect robot motion~\cite{optical_sensor2,magnetic_sensor1}.
However, many of these sensors are high-cost, and demand specific environmental conditions, such as electric fields, optical fibers, or magnetic fields~\cite{softrobot_overview}. 
Moreover, in minimally invasive surgery, attaching these sensors to small robotic instruments may pose potential safety risk. 

Recently, image-based methods have gained traction owing to their ability to eliminate the need of physically attaching sensors to robots. 
Among these, keypoint-based methods are commonly used for rigid robot state estimation.
These methods employ deep learning models to identify the 2D positions of predefined 3D points. 
Subsequently, state parameters are derived by establishing 2D-3D correspondences using the algorithms such as Perspective-n-Point (PnP).
For instance, PVNet~\cite{PVNet} first segments the input image using a convolutional neural networks (CNN)~\cite{dai2023automatic}, and then computes the pixel-wise vectors from the segmentation mask. 
In a similar vein, DREAM~\cite{DREAM} estimates robot state by applying VGG~\cite{yan2015hd} to extract 2D belief maps, which are then used to regress the camera-to-robot state using these keypoints, forward kinematics, and camera intrinsics. 
Following this pipeline, a graph-based method~\cite{Graph-based} introduces graph representation to refine the location of keypoints in surgical robot state estimation.
However, detecting keypoints on flexible robots is challenging due to the absence of clearly distinguishable robot joints.
To address this, some end-to-end frameworks have been proposed to directly regress state from extracted features. 
PoseNet~\cite{PoseNet}, for example, uses a single RGB image as input to estimate the object's state. 
Single-Shot~\cite{Single-shot} extends a SSD-like architecture to predict surgical robot state parameters through state maps. 
Likewise, SimPS-Net~\cite{SimPS} combines segmentation and state estimation by adding a state regression branch to the head of Mask-RCNN. 
Previous studies have shown that direct regression can outperform keypoint-based methods in camera calibration tasks~\cite{CoRL}.
However, these regression-based methods typically provide a single best estimate of the robot state without indicating uncertainty, which could lead to severe consequences if the state is poorly estimated. 
In contrast, our approach generates a distribution of rotations for the flexible instruments, considering both the robot's state and associated uncertainty, which allows for a robust and reliable outcome.

\subsection{Surgical Scene Depth Estimation}

Several studies have explored supervised learning methods for depth estimation~\cite{mahmood2018deep,allan2021stereo}. However, obtaining labeled data in clinical scenarios for training depth estimation networks remains challenging.
Recently, the emergence of vision foundation models has significantly advanced monocular depth estimation. Models such as Depth Anything~\cite{yang2024depth} and Metric3Dv2~\cite{hu2024metric3d} have demonstrated remarkable generalization capabilities through large-scale pre-training on diverse datasets. Following this trend, several works have adapted these foundation models to endoscopic imaging. Self-supervised approaches including EndoUFM~\cite{yao2025endoufm}, EndoDAV~\cite{zhou2025endodav}, EndoDAC~\cite{cui2024endodac}, and DARES~\cite{sheikh2024dares} fine-tune foundation models using photometric consistency losses, which assume static endoscopic environments and utilize warping-based constraints between consecutive frames. In contrast, EndoOmni~\cite{tian2024endoomni} employs supervised learning with ground-truth depth annotations. However, our robotic endoluminal surgery scenario presents unique challenges: the presence of flexible instrument motion and dynamic tissue deformation violates the static scene assumption of self-supervised methods, while obtaining ground-truth depth for these complex scenes is particularly difficult.

Furthermore, photometric stereo methods have been developed to reconstruct surfaces from multiple images under varying illumination conditions~\cite{yang2022ps}. These approaches model scene lighting to estimate depth through optimization frameworks~\cite{santo2020deep}. LightNeuS~\cite{batlle2023lightneus} exploits the inverse-square relationship between illumination and distance in co-axial endoscopic lighting for multi-view 3D surface reconstruction of static tissue. For monocular scenarios, LightDepth~\cite{rodriguez2023lightdepth} incorporates illumination attenuation with differentiable rendering through self-supervised loss functions for endoscopic depth estimation.
However, these photometric methods primarily focus on tissue surface reconstruction and do not account for articulated flexible instruments, which introduce complex geometry deformations. In contrast, our approach performs learning-based monocular depth estimation in dynamic robotic endoluminal environments containing both flexible instruments and deformable tissue.

\begin{figure*}[t]
\centering
\includegraphics[width = 0.95\hsize]{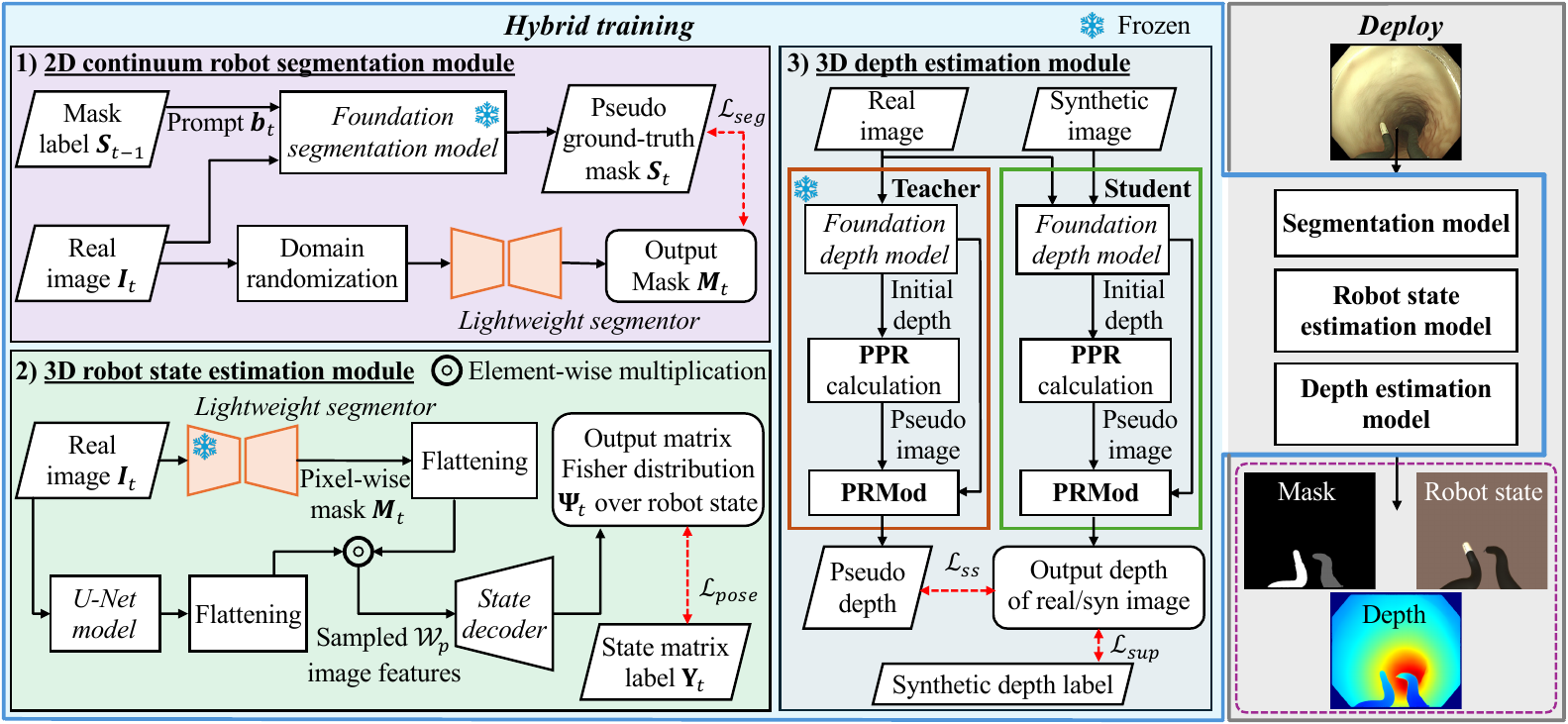}
\caption{Overview of the proposed image-based perception framework for continuum robotic endoluminal surgery. The framework consists of three modules in its hybrid training: 1) 2D continuum robot segmentation module for flexible instruments recognition; 2) 3D robot state estimation module for flexible instrument shape calculation ; and 3) 3D depth estimation module for geometric information provision about the entire endoscopic scene. Monocular endoscopic images serve as the sole input for the perception framework. After training, the framework is deployed on a novel continuum robotic system.}
\label{fig:method}
\vspace{-0.3cm}
\end{figure*} 

\subsection{Simulator for Surgical Robotic System}
Several surgical robotic simulators have been developed, which primarily focus on rigid robotic instruments in virtual surgical environments. SuRoL~\cite{xu2021surrol} is a dVRK compatible simulation platform for surgical robot learning, which focuses on the rigid instrument manipulation and robotic motion collection. Similarly, AMBF~\cite{munawar2019real} offers a 3D virtual environment about robotic laparoscopy. It leverages the front-end description format to simulate the multi-body rigid surgical instruments. Moreover, the synthetic data generated by AMBF is primarily aimed at robot action learning rather than perception learning, which necessitates a realistic background and accurate 3D structure. 
Other platforms, like dVRL~\cite{richter2019open} and UnityFlexML~\cite{tagliabue2020soft}, are designed for reinforcement learning with dVRK, but struggle to produce realistic simulated data for perception models. 
NVIDIA Isaac for Healthcare is an end-to-end platform that accelerates medical robotics development through simulation, synthetic data generation, and real-time deployment. Building upon this platform, SonoGym~\cite{ao2025sonogym} develops a simulation environment for robotic ultrasound tasks, offering synthetic ultrasound image generation. Similarly, robot-assisted surgical suturing, originally developed in AMBF, has been adapted for Isaac Sim~\cite{kim2025surgical}.
Recently, simulation environments have been developed to generate synthetic data for endoscopy-related perception tasks. For instance, VR-Caps~\cite{incetan2021vr} is a simulation platform for capsule endoscopy operations, which can generate fully labeled and realistic synthetic data for data-driven perception algorithms. It simulates a range of normal and abnormal tissues conditions as well as different organ types. However, VR-Caps does not model complex flexible instruments, thus limiting its ability to generate endoscopic scene data with continuum robot motion. In contrast, our physically-realistic simulator models flexible robotic instruments and further simulates diverse 3D organ models with realistic textures, which allows us to gain large amounts of synthetic data for developing learning-based perception algorithms. Additionally, this simulator can be used to monitor the 3D state of the flexible instruments after connecting the simulation to the real robotic system. 

\section{Endoscopic-image-based Perception Framework}
\label{sec:software}

\subsection{Overview of the Image-based Perception Framework}

To provide surgeons with cognitive assistance in controlling the continuum robotic system, we design an image-based perception system that incorporates a set of advanced learning-based algorithms, as illustrated in Fig.~\ref{fig:method}. This system focuses on recognizing flexible robotic instruments, monitoring their states (the 3D shape of the robotic instruments), and performing depth estimation of both the flexible instruments and dynamic surgical scenes. Specifically, given a real-time endoscopic video stream, our image-based perception system can automatically segment robotic instruments using an annotation-efficient segmentation model. Based on this segmentation, the degrees of freedom of the flexible robotic instruments can be inferred through an image-based robot state estimation model. Additionally, depth information is obtained by a monocular depth estimation foundation model, which infers the 3D deviation between instruments and soft tissues. Consequently, this model significantly enhances the surgeon's perception of depth, thereby improving their clinical decision-makings during surgical procedures.

\subsection{2D Annotation-efficient Continuum Robot Segmentation}
\label{sec:segmentation}
Accurate segmentation of flexible surgical instruments is a critical challenge in image-based perception.
To fully leverage the real-world unlabeled data with diverse surgical backgrounds for robust segmentation, we propose an annotation-efficient segmentation module for flexible instruments.
The core idea is to utilize recent powerful vision foundation models, which are pre-trained on large-scale datasets, for effective segmentation.
However, directly applying these foundation models can be inefficient in practice.
To overcome this limitation, we introduce a knowledge distillation process that transfers the expertise of the foundation model to a lightweight network, ensuring efficient and effective segmentation~\cite{hakkinen2024medical}.

The overall module is illustrated in Fig.~\ref{fig:method}. 
Specifically, we build our module on top of the foundation segmentation model named Segment Anything Model (SAM)~\cite{sam}.
Instead of requiring extensive annotations for instrument contours in every image, we only need to customize simple bounding boxes $\bm{b}^l_0$ and $\bm{b}^r_0$ for left and right instruments in the initial frame $\bm{I}_0$ of each image sequence.
Using this location information as a segmentation prompt, we apply SAM to effectively segment the left and right instruments within the specified bounding box, denoted as $\bm{S}^l_t$ and $\bm{S}^r_t$ at time $t$, respectively. 
The bounding box area for frame $\bm{I}_t$ can be automatically adjusted using the segmentation results from frame $\bm{I}_{t-1}$. For the left flexible instrument, the updated bounding box $\bm{b}^l_t$ at frame $\bm{I}_t$ is given by:
\begin{equation}
\begin{aligned}
    & \bm{b}^l_t = [\text{max}(0,\bm{S}^l_{t-1}(lm)-\varepsilon), \quad \text{min}(W, \bm{S}^l_{t-1}(rm)+\varepsilon), \quad \\ 
    & \text{max}(0, \bm{S}^l_{t-1}(tm)-\varepsilon), \quad \text{min}(H, \bm{S}^l_{t-1}(bm)+\varepsilon)].
\end{aligned}
\end{equation}
where $\bm{S}^l_{t-1}(lm)$, $\bm{S}^l_{t-1}(rm)$, $\bm{S}^l_{t-1}(tm)$, and $\bm{S}^l_{t-1}(bm)$ represent the leftmost, rightmost, topmost, and bottommost pixel coordinates in the binary mask $\bm{S}^l_{t-1}$, respectively. Here, $\varepsilon$ is a constant for bounding box padding, while $W$ and $H$ are the width and height of the frame image.
This process is highly automatic, eliminating the need for human annotation. 

For the lightweight segmentation network, we adopt a U-Net-based encoder-decoder architecture as our student model.
Utilizing the semantic labels $\bm{S}^l_t$ and $\bm{S}^r_t$ predicted by SAM as ground truth, we supervise the learning of the lightweight segmentor using negative log likelihood (NLL) loss:
\begin{equation}
    \mathcal{L}_{seg}=-\frac{1}{H \times W} \sum_{u=1}^{H} \sum_{v=1}^{W} 
    \sum_{k=1}^{K}
    y_{u,v,k} \text{log}(p_{u,v,k}).
\end{equation}
where $\left(u,v\right)$ denotes the pixel coordinate in the image, $K=2$ is the number of classes, and $y_{u,v,k}$ and $p_{u,v,k}$ are the output probability of class $k$ from SAM and lightweight segmentor, respectively.
The segmentation mask $\bm{M}$ can then be derived by assigning each pixel $\bm{m}$ the class $k$ with the highest probability: $\bm{m}_{u,v}=\arg\mathop{\max}\limits_{k} \; p_{u,v,k}$.
This enables the lightweight network to effectively learn from the detailed information provided by the foundation model, capturing important spatial and semantic features. 
To further enhance segmentation robustness in the endoscopic scenarios, we propose a semantic-guided data augmentation technique. 
We leverage the binary masks from the foundation model to segment the instruments areas and paste the cropped segments onto diverse endoscopic backgrounds randomly sampled from the public Endoscopy Artefact Detection (EAD) dataset~\cite{ead2}. 
Subsequently, we apply data augmentation strategies to adjust the image brightness, contrast, saturation, and hue, as well as to add random Gaussian noise.
This approach significantly improves the robustness of the segmentation model across various endoscopic scenes.

\subsection{3D Image-based Continuum Robot State Estimation}

With the segmentation of two flexible instruments, we further estimate the state of the continuum robots in 3D space. An overview of the proposed module for flexible robot state estimation is depicted in Fig.~\ref{fig:method}. 
First, we extract the endoscopic image features with an encoder-decoder architecture built on U-Net. 
Next, we sample $\mathcal{W}_p$ pixel-wise robot features based on the corresponding segmentation mask of the flexible instruments. 
A straightforward approach to resolve the state is to directly regress each parameter from the image features.
However, it is widely recognized that neural network outputs are unconstrained values that span $\mathbb{R}^N$, while the state parameters exist within a rotation space, a non-linear and closed manifold. 
This discrepancy presents a challenge in defining a meaningful loss function, as it can lead to disconnected local minima that are poorly defined~\cite{CoRL}.
Traditional representations such as Euler angles, rotation vectors, and quaternions cannot serve as optimization targets unless constraints are added to the network or the loss function~\cite{mahendran20173d}.
To address this issue, we adopt matrix Fisher distribution to model the state parameters probabilistically, which enables a continuous learning space for state regression~\cite{ProbabilityOrien}.
Specifically, the matrix Fisher distribution defines a probability density function over the rotation matrix $\mathbf{R} \in \mathbb{R}^{3 \times 3}$ as follows:
\begin{equation}
p(\mathbf{R}) = \mathcal{M}(\mathbf{R}; \mathbf{\Psi}) = \frac{1}{n(\mathbf{\Psi})}\exp(\text{tr}(\mathbf{\Psi}^\top\mathbf{R})).
\end{equation}
where $\mathbf{\Psi} \in \mathbb{R}^{3 \times 3}$ represents the distribution parameters, $\mathcal{M}(\mathbf{R};\mathbf{\Psi})$ is a probability distribution over $\textit{SO}(3)$ for rotation matrices, $n(\mathbf{\Psi})$ is a normalizing constant, and $\text{tr}$ denotes the trace of the matrix.
The regressed distribution parameters $\mathbf{\Psi}$ not only facilitate the recovery of the robot state values but also provide corresponding uncertainty values that indicate the confidence of the prediction. 
During the training phase, we employ the NLL loss to train the probabilistic model, which is defined as follows:
\begin{equation}
\mathcal{L}_{pose} = -\text{log}(\mathcal{M}(\mathbf{Y}; \mathbf{\Psi})).
\end{equation}
where $\mathbf{Y}$ represents the ground-truth state matrix. 
During the inference phase, given the model output $\mathbf{\Psi}$, we can obtain the mode and dispersion of the distribution by performing singular value decomposition (SVD) on $\mathbf{\Psi}$, denoted as $\mathbf{\Psi} = \mathbf{USV}^T$. 
The state value can be derived from $\mathbf{U}$ and $\mathbf{V}$, while the singular values in $\mathbf{S}$ indicate the concentration of the distribution, providing insights into the uncertainty of the predicted flexible robot states.

As shown in Fig.~\ref{fig:simulator}(b), our continuum robot state is characterized by four parameters. The proposed state estimation module predicts a distinct matrix Fisher distribution for each parameter, explicitly modeling their uncertainties. By integrating these four inferred parameters, the 3D shape of the flexible instrument can be fully reconstructed.

\subsection{3D Depth Estimation in Robotic Endoluminal Scenes}
Building on robot state estimation, we compute the 3D shape of flexible robotic instruments in real-time. Furthermore, understanding the entire 3D geometry of the endoscopic scene, including the tissue structure and the instruments' depth, is crucial. Therefore, we propose a monocular depth estimation module to automatically perceive and reconstruct the surgical environment in 3D. Current efforts primarily focus on understanding the surface structures of internal organs from endoscopic videos. They heavily rely on learning priors from synthetic training data. However, there is currently no synthetic dataset available that simulates flexible robots moving within internal organs. Additionally, due to the confined spaces of the endoluminal surgery, collecting clinical data with ground truth depth information is impractical.
To address this issue, we first synthesize a substantial amount of training data using our synthetic data generation framework (described in Section~\ref{sec:hardware}). The simulated flexible instruments in these datasets closely resemble the configurations of real flexible robots illustrated in Fig.~\ref{fig:simulator}(c), and the artificial surgical scenes are highly realistic. Next, to train the depth model using clinical data, we analytically compute the illumination model of the endoscope to design a self-supervised loss function. 
We then implement a teacher-student training strategy, as shown in Fig.~\ref{fig:method}, to facilitate the synthetic-to-real transfer learning. This approach involves training a depth estimation model, consisting of a foundation monocular depth model followed by an improvement module, on both the simulated data with supervision and real clinical data with the proposed self-supervision. The teacher model further guides the learning of the student network on real unlabeled surgical data through the self-supervised loss function.

\subsubsection{Illumination Modeling of Monocular Endoscope}

\begin{figure}[t]
\centering
\includegraphics[width = 0.95\hsize]{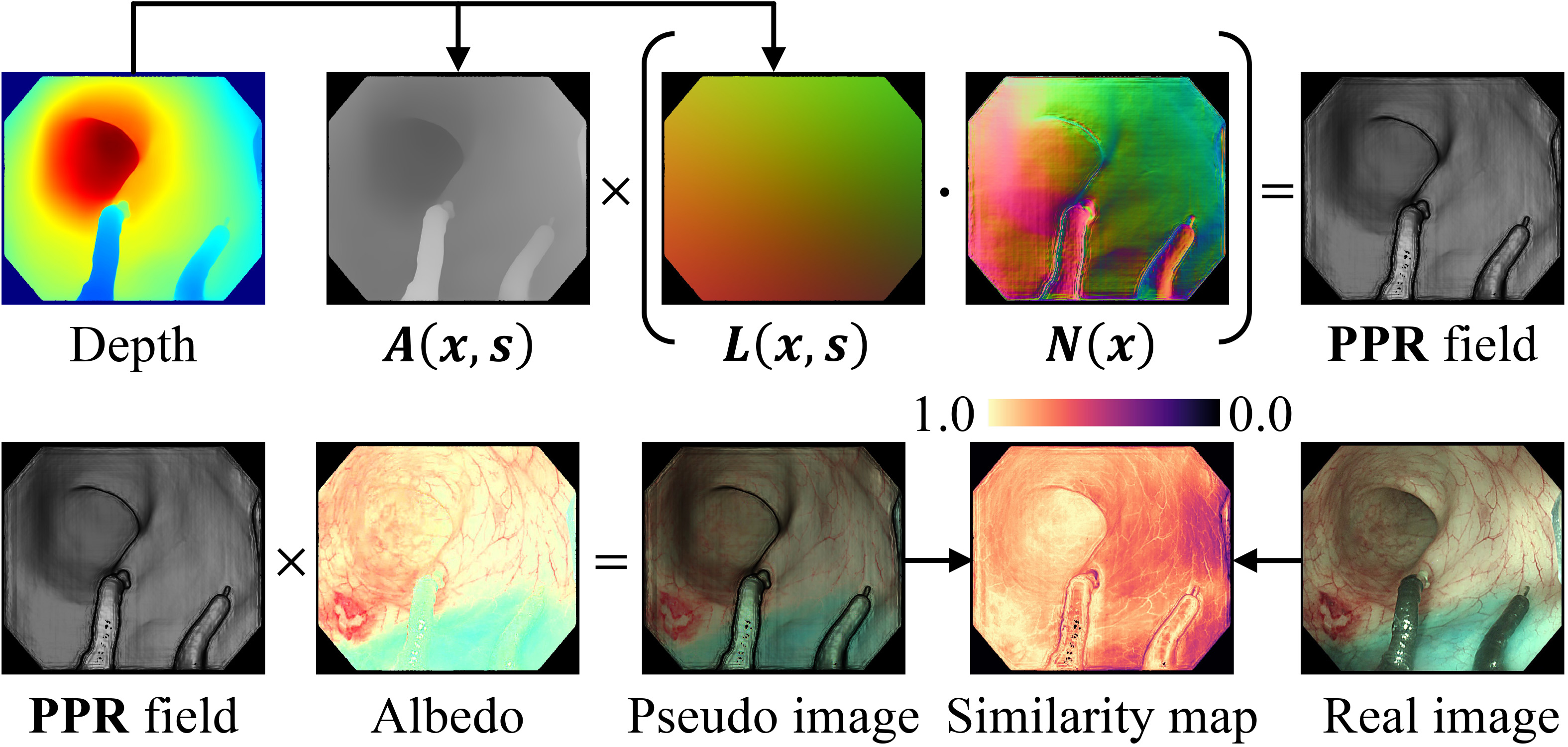}
\caption{Calculation of the Per-Pixel Rendering ($\textbf{PPR}$) field from the depth estimation, which is the process of \textbf{PPR} calculation illustrated in Fig.~\ref{fig:method}. The $\textbf{PPR}$ field shows a strong correlation with the corresponding input endoscopic image. Arrows from Depth indicate how depth is transformed into attenuation and light direction. Additionally, arrows between Pseudo and Real images represent the correlation computation, with the correlation values normalized to a range of 0 to 1.}
\label{fig:ppr}
\vspace{-0.35cm}
\end{figure} 

Current depth estimation methods primarily rely on learning geometric and semantic priors from their training datasets. However, these methods often overlook critical photometric information that can be derived from the co-located light source and endoscope. Surfaces of internal organs that are closer to the endoscope and facing it receive more incident illumination than those that are farther away or oriented in the opposite direction. 
Following the approach in~\cite{modrzejewski2020light}, we model the scene illumination as emanating from a single point light source in the camera reference frame. 
Therefore, the general rendering equation for each pixel $\bm{p} \in \mathbb{R}^2$ in endoscopic image $\bm{I} \in \mathbb{R}^{H \times W \times C}$, where $(H,W)$ denotes the image resolution and $C$ is the number of channels, is expressed as:
\begin{equation}
\label{equ:ill_render}
\bm{I}(\bm{p}) = f_c(\bm{p}) \; f_r(\rho(\bm{x}), \bm\sigma(\bm{x}, \bm{s}), \bm{N}(\bm{x})).
\end{equation}
where $\bm{x} \in \mathbb{R}^3$ is the 3D point corresponding to pixel $\bm{p}$, $\bm{s} \in \mathbb{R}^3$ is the 3D position of the point light, $\bm\sigma(\bm{x}, \bm{s})$ is the light vector relating to the light received by the the surface at $\bm{x}$, while $\rho(\bm{x})$ and $\bm{N}(\bm{x})$ represent the albedo and normal at $\bm{x}$, respectively. The function $f_r(\cdot)$ describes the reflectance model of the material, and $f_c(\cdot)$ is the camera response function, which is often assumed to be constant across different views~\cite{langguth2016shading}. 
Given that colon surfaces exhibit minimal variance in albedo, we set $\rho(\bm{x})$ to a constant value by converting the RGB image to HSV space and extracting the hue and saturation components, with the value channel set to 100\%.  
For lambertian reflectance~\cite{szeliski2022computer}, we have:
\begin{equation}
f_r(\rho(\bm{x}), \bm\sigma(\bm{x}, \bm{s}), \bm{N}(\bm{x})) = \rho(\bm{x}) \cdot\bm\sigma(\bm{x}, \bm{s}) \cdot \bm{N}(\bm{x}).
\end{equation}
which simplifies the rendering equation to:
\begin{equation}
\bm{I}(\bm{p}) \sim \bm\sigma(\bm{x}, \bm{s}) \cdot \bm{N}(\bm{x}).
\end{equation}
Next, we model the endoscope light as a point source $\bm{s}$, allowing us to express the light vector $\bm\sigma(\bm{x}, \bm{s})$ as:
\begin{equation}
\bm\sigma(\bm{x}, \bm{s})= \sigma_0 \; \bm{A}(\bm{x}, \bm{s}) \; \bm{L}(\bm{x}, \bm{s}).
\end{equation}
where
\begin{equation}
\bm{A}(\bm{x}, \bm{s}) = \frac{1}{\left\|\bm{x}-\bm{s}\right\|^2}, \; \bm{L}(\bm{x}, \bm{s}) = \frac{\bm{x} - \bm{s}}{ \left\|\bm{x}-\bm{s} \right\|}.
\label{equ:L_A}
\end{equation}
where $\sigma_0$ represents the maximum radiance, $\bm{A}$ is the attenuation factor map, and $\bm{L}$ denotes the lighting direction. Thus, we can rewrite the rendering equation as:
\begin{equation}
\bm{I}(\bm{p}) \sim  \underbrace{\bm{A}(\bm{x}, \bm{s})\left(\bm{L}(\bm{x}, \bm{s}) \cdot \bm{N}(\bm{x})\right)}_{\textbf{PPR}}.
\label{equ:ppr}
\end{equation}
This introduces a new representation, the per-pixel rendering ($\textbf{PPR}$) field, as illustrated in Fig.~\ref{fig:ppr}, which indicates how light emitted from the endoscope is reflected by the tissue surface. The $\textbf{PPR}$ field combines the light direction $\bm{L}$, attenuation factor $\bm{A}$, and the surface normal $\bm{N}$ to capture the effects of incident lighting on both the tissue surface and flexible robots.

Using the $\textbf{PPR}$ representation, we propose a supervised loss function for training on synthetic data and a self-supervised loss function for training on clinical data.
The supervised PPR loss is defined as: 
\begin{equation}
\begin{aligned}
\mathcal{L}_{sup} = & \frac{1}{H\times W}\sum_{u=1}^{H}\sum_{v=1}^{W} \\
& {\left(\bm{E}\left(u,v\right) \left(\textbf{PPR}^{\ast}\left(u,v\right) - \textbf{PPR}_{gt}\left(u,v\right)\right)^2\right)}.
\end{aligned}
\label{equ:l_sup}
\end{equation}
Here, $\bm{E}$ is a mask used to filter specularities, defined as $\bm{E}(u,v) = 1, if~\bm{I}_g < 0.98$, where $\bm{I}_g$ is the intensity map of the image. The ground-truth, $\textbf{PPR}_{gt}$, is computed using the ground-truth depth, while the estimation, $\textbf{PPR}^{\ast}$, is derived from the final depth estimate. The supervised PPR loss is complementary to the supervised depth loss due to the inherent smoothness and low-frequency bias in depth maps compared to higher-frequency information in the $\textbf{PPR}$ field. 
The self-supervised PPR loss is described as follows:
\begin{equation}
    \mathcal{L}_{ss} = 1 - f_{corr}\left(\bm{E} \cdot \textbf{PPR}^{\ast}, \bm{E} \cdot \bm{I}_g\right).
\label{equ:l_ss}
\end{equation}
where the correlation is given by:
\begin{equation}
\begin{aligned}
f_{corr}\left(\textbf{PPR}^{\ast}, \bm{I}_g\right) & =  \\
& \frac{\sum_{u=1}^{H}\sum_{v=1}^{W}(\textbf{PPR}^{\ast} - \bar{\textbf{PPR}^{\ast}}) \cdot (\bm{I}_g - \bar{\bm{I}_g})}{\sqrt{ \sum_{uv}(\textbf{PPR}^{\ast} - \bar{\textbf{PPR}^{\ast}})^2 \cdot \sum_{uv}(\bm{I}_g - \bar{\bm{I}_g})^2}}.
\end{aligned}
\label{equ:corr}
\end{equation}
where $\bar{\textbf{PPR}^{\ast}}$ and $\bar{\bm{I}_g}$ denote the scalar mean intensities of $\textbf{PPR}^{\ast}$ and $\bm{I}_g$.
As described in (\ref{equ:ppr}),  the $\textbf{PPR}$ is strongly correlated with the image intensity field, except in regions of strong specularity. Fig.~\ref{fig:ppr} also illustrates this strong relationship between the $\textbf{PPR}$ and the input endoscopic image. This high correlation suggests that the model's predictions for $\textbf{PPR}$ align closely with changes in the image intensity field, indicating consistent and accurate modeling. 
The self-supervised loss function will enable effective training on real clinical data where ground-truth depth information is unavailable.

\subsubsection{Monocular Depth Estimation Model}

\begin{figure}[t]
\centering
\includegraphics[width = 0.95\hsize]{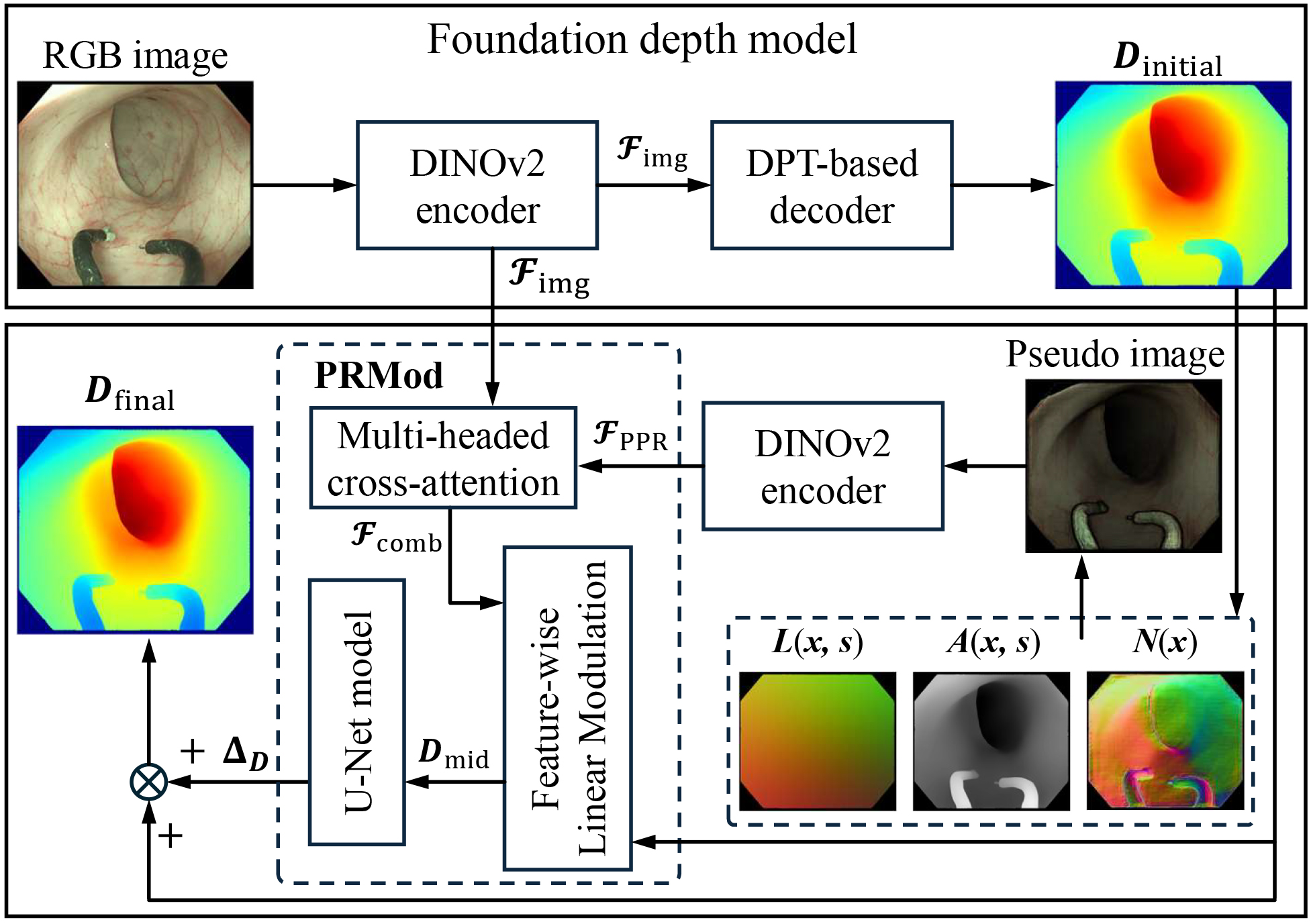}
\caption{Monocular depth estimation model for continuum robotic endoluminal surgical scene.}
\label{fig:depth_model}
\vspace{-0.35cm}
\end{figure} 

Utilizing the aforementioned loss functions and the synthetic-to-real transfer learning strategy, we train a depth estimation model in Fig.~\ref{fig:depth_model} that combines a foundation monocular depth estimation network with an improvement module, denoted as $\textbf{PRMod}$. Given a monocular endoscopic image of the flexible robotic endoluminal surgery scene, we first extract the image features $\bm{\mathcal{F}}_{\text{img}}$ using a DINOv2 encoder~\cite{oquab2023dinov2}. The initial depth estimation $\bm{D}_{\text{initial}}$ is then obtained from a DPT-based decoder~\cite{ranftl2021vision}. Further, we calculate the lighting direction $\bm{L}(\bm{x}, \bm{s})$, attenuation map $\bm{A}(\bm{x}, \bm{s})$, and normal map $\bm{N}(\bm{x})$ from the initial depth according to (\ref{equ:L_A}). By combining this information with the proxy albedo, we synthesize a pseudo-RGB image from the initial depth estimates, as illustrated in Fig.~\ref{fig:ppr}. Similarly, we extract the feature $\bm{\mathcal{F}}_{\text{PPR}}$ from the pseudo-RGB image by the DINOv2 encoder. After that, the proposed improvement module $\textbf{PRMod}$ takes $\bm{\mathcal{F}}_{\text{img}}$ and $\bm{\mathcal{F}}_{\text{PPR}}$ to fine-tune the depth $\bm{D}_{\text{initial}}$. Specifically, we first combine these feature maps through multi-headed cross-attention~\cite{vaswani2017attention}, resulting in a synthesized feature $\bm{\mathcal{F}}_{\text{comb}}$. 
Subsequently, the initial depth is adapted to depth $\bm{D}_{\text{mid}} = f_\alpha(\bm{\mathcal{F}}_{\text{comb}}) \odot \bm{D}_{\text{initial}} + f_\beta(\bm{\mathcal{F}}_{\text{comb}})$ via feature-wise linear modulation (FiLM)~\cite{perez2018film}, where $f_\alpha(\cdot), f_\beta(\cdot)$ are scale and shift functions implemented as a linear transformation $f_l(\cdot)$ of $\bm{\mathcal{F}}_{\text{comb}}$ (i.e., $f_\alpha = f_l^{(0)}(\bm{\mathcal{F}}_{\text{comb}})$, $f_\beta = f_l^{(1)}(\bm{\mathcal{F}}_{\text{comb}})$), and $\odot$ denotes element-wise multiplication, following the standard FiLM formulation.
Finally, the depth map $\bm{D}_{\text{mid}}$ is input to a four-layer U-Net to calculate the refinement value $\Delta_{\bm{D}}$, which is then added to $\bm{D}_{\text{initial}}$ to yield final depth $\bm{D}_{\text{final}}$.

\subsubsection{Synthetic-to-Real Training of Depth Estimation Model}

\begin{algorithm}[t]
\caption{Synthetic-to-Real Transfer Learning for Continuum Robot Surgical Scene Data}
\label{alg:algorithm-1}
\begin{algorithmic}[1]
\Require labeled synthetic dataset: $\mathcal{T}_L$, unlabeled clinical dataset: $\mathcal{T}_U$.
\State Initialize teacher and student networks ($\mathcal{N}_t$, $\mathcal{N}_s$) with the same architecture.
\State Train $\mathcal{N}_t$ on $\mathcal{T}_L$ with (\ref{equ:l_sup}) using loss:
$\mathcal{L}=\lambda_{ssi} \mathcal{L}_{ssi} + \lambda_{re} \mathcal{L}_{re} + \lambda_{vn} \mathcal{L}_{vn} + \lambda_{sup} \mathcal{L}_{sup}$
\State Freeze $\mathcal{N}_t$ to prevent further optimization.
\State Prepare hybrid data $\mathcal{T}_H$ combining $\mathcal{T}_L$ and $\mathcal{T}_U$.
\For{each batch $i$ in $\mathcal{T}_H$}
    \If{$i$ is from $\mathcal{T}_L$}
       \State Train $\mathcal{N}_s$ on $\mathcal{T}_L$ with (\ref{equ:l_sup}) using loss:
       $\mathcal{L}=\lambda_{ssi} \mathcal{L}_{ssi} + \lambda_{re} \mathcal{L}_{re} + \lambda_{vn} \mathcal{L}_{vn} + \lambda_{sup} \mathcal{L}_{sup}$
    \EndIf
    \If{$i$ is from $\mathcal{T}_U$}
       \State Train $\mathcal{N}_s$ on $\mathcal{T}_U$ with (\ref{equ:l_ss}) and (\ref{equ:corr}) using loss:
       $\mathcal{L}=\lambda_{ssi} \mathcal{L}_{ssi^*} + \lambda_{re} \mathcal{L}_{re^*} + \lambda_{vn} \mathcal{L}_{vn^*} + \lambda_{ss} \mathcal{L}_{ss}$
       \State Calculate $\mathcal{L}_{ssi^*}$, $\mathcal{L}_{re^*}$, and $\mathcal{L}_{vn^*}$ using pseudo supervision from $\mathcal{N}_t$.
    \EndIf
\EndFor
\end{algorithmic}
\end{algorithm}

We initially train the proposed network on our synthetic continuum robot surgical scene data with supervision. Specifically, we employ scale-shift invariant ($\mathcal{L}_{ssi}$), depth regularization ($\mathcal{L}_{re}$), and virtual-normal ($\mathcal{L}_{vn}$) losses as described in ~\cite{ranftl2020towards}, along with our proposed $\mathcal{L}_{sup}$ PPR loss. While this network performs excellently on the synthetic dataset, we observe that it often produces errors and artifacts when applied to real clinical data. This discrepancy is a common issue for any neural network trained on synthetic data and tested on real data, known as the synthetic-to-real gap. To mitigate this problem, we propose teacher-student learning approach, where a student network is further trained on real data with self-supervision, guided by the teacher network. 
This technique is similar to ~\cite{yang2024depth} that utilizes large-scale data and trains using a self-supervised loss approach with semantic losses.
Our synthetic-to-real transfer learning leverages illumination information to train on 25000 real clinical images using self-supervision.

Our student network has the same architecture as the teacher network. It is trained using both labeled synthetic dataset ($\mathcal{T}_L$) and unlabeled clinical dataset ($\mathcal{T}_U$) in an interleaved manner, which enhances stability during training. When using labeled synthetic data we apply the same supervised loss functions as the teacher. However, when training on unlabeled real data, we lack ground-truth depth. Instead, we utilize the proposed self-supervised loss function $\mathcal{L}_{ss}$ for training. However, self-supervision alone is often insufficient for producing accurate depth estimates due to inherent ambiguities in training. Therefore, we employ guidance from the teacher network in terms of pseudo-supervision, where the output of the teacher network for a clinical image is used as a pseudo ground-truth for computing the supervised loss functions over depth map and $\textbf{PPR}$ field. Our approach, utilizing teacher-student training on both synthetic data $\mathcal{T}_L$ and clinical data $\mathcal{T}_U$, is summarized in algorithm~\ref{alg:algorithm-1}.
\section{Simulation of Flexible Robotic Instruments in Endoluminal Surgery} \label{sec:hardware}

\subsection{Overview of the Physically-realistic Simulation}

\begin{figure}[t]
\centering
\includegraphics[width = 0.95\hsize]{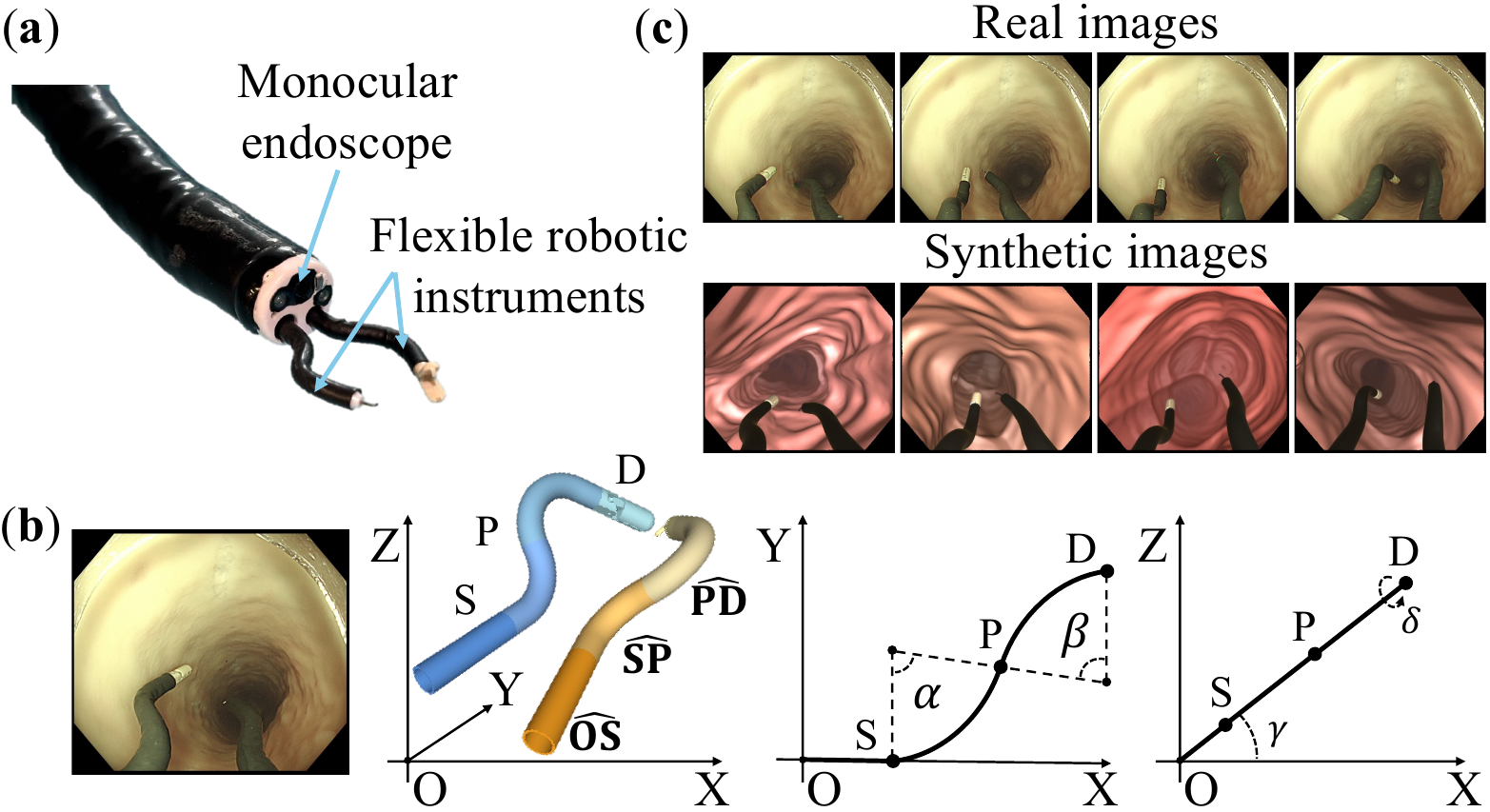}
\caption{Physically-realistic simulation of continuum robotic system for endoluminal surgery. (a) Setup of the continuum robotic system. (b) Configuration of flexible instruments approximated by curve segments in camera coordinate system. The flexible robot consists of three segments: Shoulder $\widehat{\textbf{OS}}$, Proximal $\widehat{\textbf{SP}}$, and Distal $\widehat{\textbf{PD}}$. $\alpha$ and $\beta$ denote internal bending angles of $\widehat{\textbf{SP}}$ and $\widehat{\textbf{PD}}$. $\gamma$ and $\delta$ represent $\widehat{\textbf{OS}}$ yaw and roll angles. (c) Comparison of real and synthetic endoscopic images. More comparison results are shown in Video 1.}
\label{fig:simulator}
\vspace{-0.2cm}
\end{figure}  

To enhance the ease, speed, and precision of endoscopic and endoluminal surgery, continuum robotic systems equipped with flexible instruments are increasingly surged. 
Given the small size of these instruments and the limited operating space at the surgical site, real-time 3D visualization of the robotic instruments is of paramount importance.
To achieve this, we have developed a physically-realistic simulator using Unity, the popular graphics engine and a real-time 3D development platform.  
This simulator not only models the flexible robotic instruments but also generates a substantial quantity of domain-specific data essential for training learning-based perception algorithms described in Section~\ref{sec:software}.
Fig.~\ref{fig:simulator}(a) illustrates the setup of the real continuum robotic system for endoluminal surgery. The simulation system, which mimics the actual robotic system, consists of two simulated flexible instruments, 3D organ models, a camera that simulates a monocular endoscope, and a light source. Additionally, a multi-modal camera is integrated into the simulator to generate  ground-truth depth and normal maps for the synthetic data. 

To model the synthetic instruments as realistically as possible, it is essential to accurately define the robot states of the real flexible instruments in configuration space. 
Unlike kinematics for traditional rigid robots, where the state of any point can be fully defined by link lengths and joint angles, we employ the piece-wise constant curvature model to describe the states of the robotic instruments.
As shown in Fig.~\ref{fig:simulator}(b), in addition to the end effector, the flexible robot is divided into three segments: Shoulder $\widehat{\textbf{OS}}$, Proximal $\widehat{\textbf{SP}}$, and Distal $\widehat{\textbf{PD}}$. Among these segments, $\widehat{\textbf{SP}}$ and $\widehat{\textbf{PD}}$ are non-rigid, deformable arcs in the $\text{XOY}$ plane, while $\widehat{\textbf{OS}}$ is a flexible segment capable of rotation in the $\text{ZOX}$ plane and insertion along the axis. We define $\alpha$ and $\beta$ as the arc angles of the proximal $\widehat{\textbf{SP}}$ and distal $\widehat{\textbf{PD}}$, respectively. Additionally, $\gamma$ and $\delta$ represent the yaw and roll of the shoulder $\widehat{\textbf{OS}}$.
Once these four state parameters $(\alpha, \beta, \gamma, \delta)$ are measured, the shape of the flexible instrument in 3D space can be determined. In our simulation, the flexible instrument also consists of three segments, each segment composed of several cylinders. The length ratios among the three segments match those of the real robot. We control the simulated flexible instrument using the four parameters, $(\alpha, \beta, \gamma, \delta)$, thus ensuring the consistency in motion between virtual and real robots. To compare the motion between synthetic and real instruments, we render 2D images seen from the virtual camera as the flexible robots move, as shown in Fig.~\ref{fig:simulator}(c). 

\subsection{Realistic Synthetic Data Generation from Simulator}
\begin{figure}[t]
\centering
\includegraphics[width = 0.95\hsize]{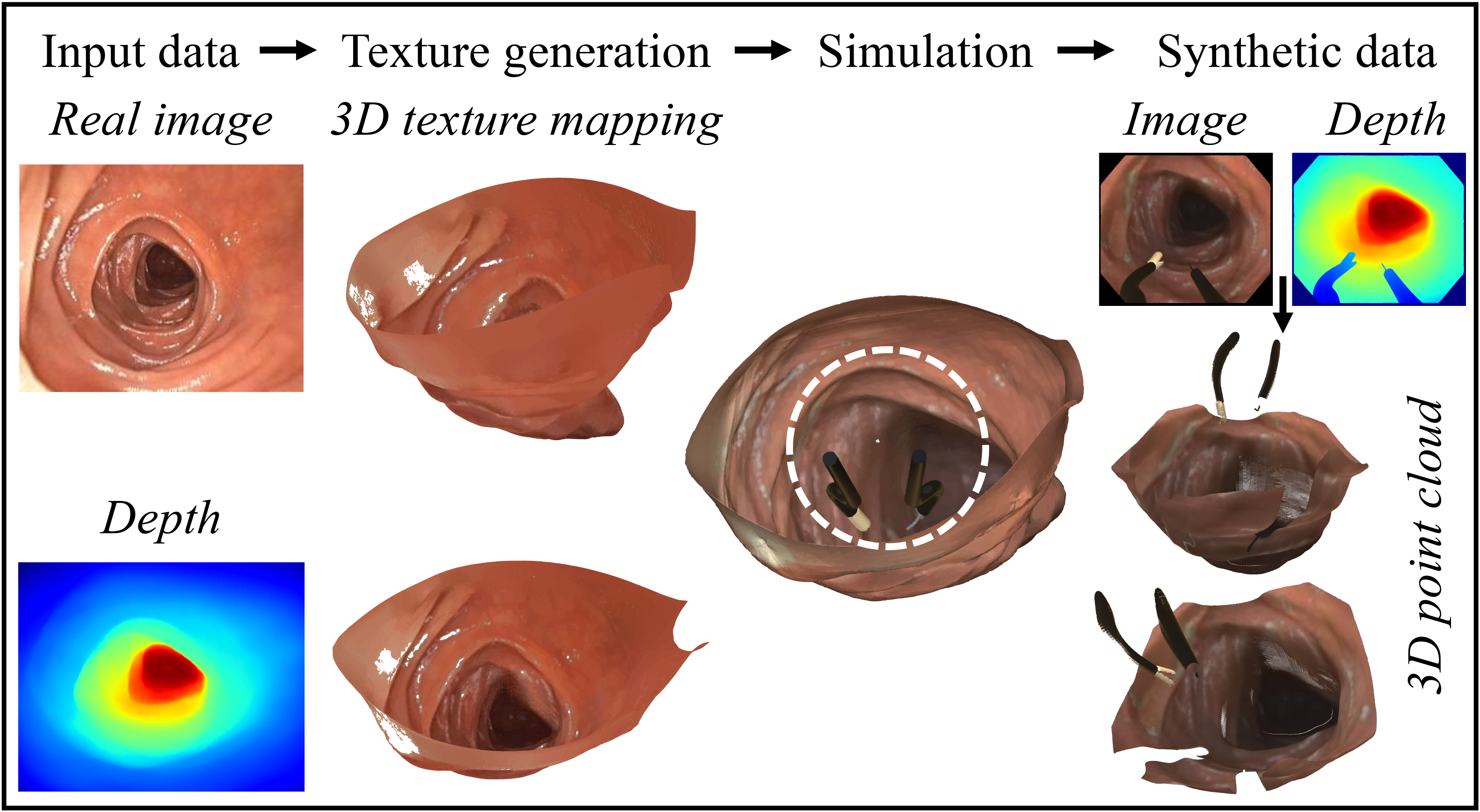}
\caption{Overview of the realistic synthetic continuum robot surgical data generation framework. More synthetic datasets can be found in Video 2.}
\label{fig:synthetic_data_platform}
\vspace{-0.2cm}
\end{figure}

To develop learning-based perception methods, we propose a framework, illustrated in Fig.~\ref{fig:synthetic_data_platform}, for generating synthetic continuum robot scene data for endoluminal surgery from the simulator. Specifically, these datasets feature simulated flexible instruments that closely match the configuration of real instruments, while the surgical scenes are highly realistic. 
We will first detail the process of generating synthetic endoscopic endoluminal scenes, followed by a description of a measurement platform designed to accurately calculate the ground-truth flexible robot states. Lastly, we will outline our strategy for recording high-quality endoscopic videos along with corresponding ground-truth depth maps.

\subsubsection{Generating Realistic Endoluminal Surgical Scene}

Our goal is to create simulated endoluminal surgical scenes with accurate 3D geometry of gastrointestinal (GI) organs and highly realistic textures. To represent the 3D geometry of GI organs precisely, we utilize the Colonoscopy 3D Video Dataset (C3VD)~\cite{bobrow2023colonoscopy}, which was acquired using a high-definition clinical colonoscope with high-fidelity physical colon models. 
The complete colon model comprises five anatomical segments: sigmoid colon, descending colon, transverse colon, ascending colon, and cecum. The dataset includes 22 short video sequences with paired ground-truth depth maps and surface normals, captured using a colonoscope rigidly mounted to an UR-3 robotic arm. Due to the rigid mounting configuration, view variation within each sequence is limited. For our simulator, we selected all five colon segments for their 3D geometric structure and extracted 12 representative frames with their ground-truth depth from the 22 video sequences.
Additionally, for real clinical data, we utilize Colon10K~\cite{ma2021colon10k}, which provides 20 calibrated colonoscopic videos. We selected one representative frame from each video. To obtain depth information for these frames, we employed a SOTA estimation foundation model~\cite{paruchuri2024leveraging} specifically designed for endoscopic scenes to generate high-accuracy pseudo depth maps.

\begin{figure}[t]
\centering
\includegraphics[width = 0.95\hsize]{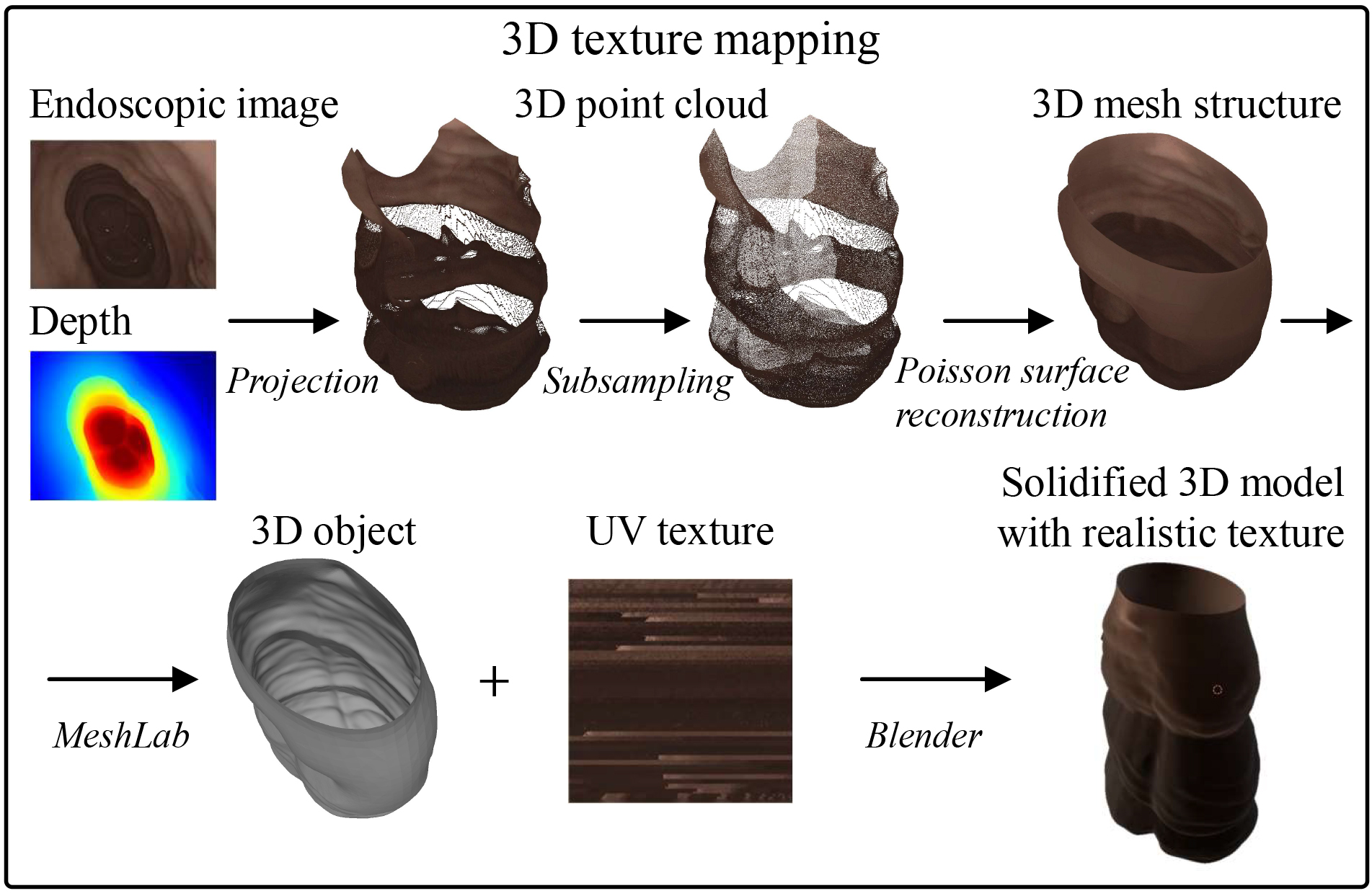}
\caption{3D texture mapping pipeline for extracting high-quality and continuous textures from endoscopic images and depth information, which is the process in Fig.~\ref{fig:synthetic_data_platform}.}
\label{fig:texture}
\vspace{-0.25cm}
\end{figure}

The next step involves textures creation for generating a synthetic realistic scene. As shown in Fig.~\ref{fig:texture}, we propose a 3D texture mapping method to extract clear, non-blurry, and continuous textures from endoscopic images and corresponding depth, in addition to traditional 2D texture mapping using single color images. 
Specifically, for each individual frame (from either C3VD or Colon10K), we project its depth map and endoscopic image into a 3D point cloud $\mathcal{Q}$ using the endoscope's intrinsics matrix $\bm{\mathcal{K}}$:
\begin{equation}
\mathcal{Q}^{(u,v)} = \bm{D}^{(u,v)} \cdot \bm{\mathcal{K}}^{-1} \cdot [u \; v \; 1]^T
\end{equation}
where $(u,v)$ denotes the pixel coordinate in depth map $\bm{D}$ and endoscopic image $\bm{I}$. This 3D point cloud includes color information extracted from the endoscopic image and normal information from the ground-truth normal map. To reduce the computational costs during rendering in the simulator, we first filter the dense point cloud using a Poisson-Disk sampling strategy. 
After this sub-sampling process, the point cloud becomes uniform while retaining local details for 3D surface reconstruction. 
We represent the points after Poisson-Disk filtering with a vector field $\Vec{V}$.  
Poisson surface reconstruction addresses the surface reconstruction problem by employing a framework of implicit functions that calculate a 3D indicator function, denoted as $\mathcal{X}$. This function assigns a value of 1 inside the model and 0 to points outside of it.
The problem thus reduces to finding the $\mathcal{X}$ such that its gradient optimally approximates the vector field $\Vec{V}$:
$\min_{\mathcal{X}} ||\nabla_{\mathcal{X}} - \Vec{V}||$. By applying the divergence operator, we can transform this into a Poisson problem: $\nabla \times (\nabla_\mathcal{X}) = \nabla \times \Vec{V} \equiv \Delta_{\mathcal{X}} = \nabla \times \Vec{V}$. After solving the Poisson problem and obtaining the 3D indicator function $\mathcal{X}$, we can directly extract the 3D surface mesh by identifying an iso-surface. From this 3D mesh structure, we extract the 3D object and its corresponding UV texture map, which are then imported into Blender for further processing. In Blender, the 3D mesh object is divided into triangular segments and projected onto the created texture uniformly using the UV mapping technique. Besides, the textured 3D model must be solidified before being imported into the simulator for synthetic data generation; if not solidified, the realistic texture cannot be rendered in the simulator.

\subsubsection{Simulated Flexible Instruments Motions}

To match the simulated robot motions in training data with those in real data, we have designed a measurement platform to calculate the ground-truth robot states of the flexible instruments. 
The platform consists of the same continuum robotic system as depicted in Fig.~\ref{fig:simulator}(a), along with two cameras positioned above and to the side of the setup. 
During the data recording phase, we control the two flexible instruments to move within porcine GI tracts while simultaneously capturing endoscopic images and corresponding motor control signals. Afterwards, we replay the recorded motor signals on the measurement platform. To ensure geometric accuracy, we manually rotate the instruments for each measurement to achieve perfect vertical alignment with the top-view camera before calculating the acr angles ($\alpha$, $\beta$) of the proximal $\widehat{\textbf{SP}}$ and distal $\widehat{\textbf{PD}}$ segments. The arch angles are computed following the flexible robot configuration analysis illustrated in Fig.~\ref{fig:simulator}(b). 
Then, the yaw $\gamma$ of the shoulder $\widehat{\textbf{OS}}$ is calculated from the side-view images, while the roll $\delta$ of the shoulder is directly obtained from the recorded motor signals.
By inputting these ground-truth robot states into the simulator, the virtual flexible instruments can accurately imitate the motion of the real flexible robotic instrument. Fig.~\ref{fig:simulator}(c) compares the flexible robotic motions in the rendered and real endoscopic images. The comparison results demonstrate that the flexible instruments in the simulation exhibit motion similar to that of the real robots when ground-truth state parameters are utilized.

\subsubsection{Data Generation}
\begin{figure}[t]
\centering
\includegraphics[width = 0.95\hsize]{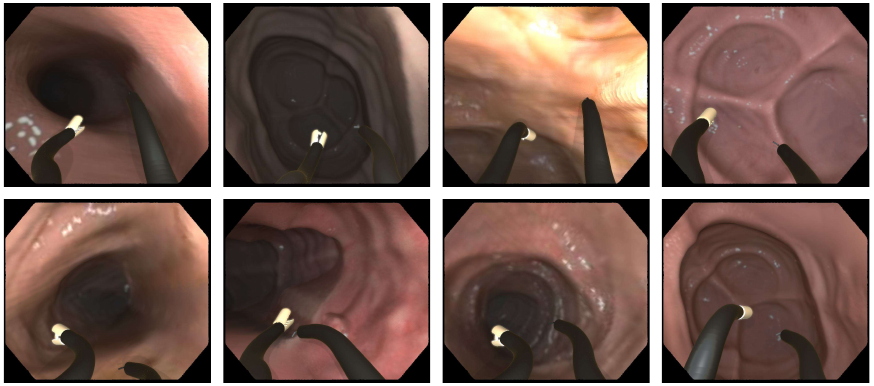}
\caption{Examples of synthetic continuum robot surgical data with multiple robot configurations and endoscope scenes.}
\label{fig:data_example}
\vspace{-0.25cm}
\end{figure}

With geometry-accurate 3D models and realistic textures of the endoscopic scene established, along with the simulated flexible robotic instruments incorporating real states parameters, we combine these elements in the simulation. A monocular camera that mimics the endoscope in our robotic system is placed within the virtual environment. 
The intrinsic matrix of the virtual camera is identical to that of the Olympus colonoscope used in our experiments, and a point light source is fixed to the camera.
To obtain ground-truth depth and normal, a multi-modal camera is also integrated into the simulator. 
As illustrated in Fig.~\ref{fig:synthetic_data_platform}, to achieve highly accurate ground-truth depth values, we decompose the normalized depth value $d \in [0, 1]$ into two 8-bit components, $d_{high}$ and $d_{low}$, using the following equations:
$d_{low} = \text{frac}(d \cdot 256)$ and $d_{high} = d - \frac{d_{low}}{256}$,
where $\text{frac}(\cdot)$ denotes the fractional part. These two values are stored in the Red and Green channels of an RGB image, respectively. During evaluation or training, the high-precision depth can be reconstructed as $d = d_{high} + \frac{d_{low}}{256}$. This approach effectively preserves depth details that would otherwise be lost due to 8-bit quantization. The scale factor for converting normalized depth to absolute depth can be determined from the known dimensions of a surgical instrument in a single reference frame and then applied consistently across all frames.
Fig.~\ref{fig:synthetic_data_platform} illustrates the 3D point cloud generated from the depth map. This dense and non-staircase point cloud accurately represents the ground-truth depth maps. With the above setting, after running the simulator, we can render endoscopic images and corresponding highly precise ground-truth labels. 
Finally, Fig.~\ref{fig:data_example} shows the diversity of the synthetic datasets with various robot configuration, endoscopic scenes, and tissue textures.

\section{Experiments}
\label{sec:module_eva}

\subsection{Experimental settings}

\subsubsection{Implementation Details}
We implement our endoscopic-image-based perception framework using PyTorch, employing Adam optimizer for training. {\textbf{Segmentation:}} For the flexible instruments lightweight segmentation model, we adopt an encoder-decoder architecture based on U-Net as our base.
During knowledge distillation, we use the SAM with a Vision Transformer Huge (ViT-Huge) model~\cite{sam} as the teacher model to supervise the U-Net.
The bounding box padding constant $\varepsilon$ is set to 20, with image dimensions $W$ and $H
$ at 1020 and 900 pixels, respectively.
To enhance model efficiency, we apply cropping and down-sampling operations to resize the images to a standardized size of $224\times224$ pixels before inputting them into the network. After processing, padding and up-sampling techniques are adopted to recover the model output to the original resolution. During training, we randomly modify 60\% of the images with various endoscope backgrounds and apply color jitter to 80\% of the images. The model is optimized with a learning rate of $2\times10^{-4}$, trained for 100 epochs with a batch size of 16 images on an NVIDIA RTX 3080 GPU. 
{\bf{Robot State:}} In flexible robot state estimation, input images are initially resized to $224\times224$ for feature extraction. Using the previous segmentation mask, we randomly sample $\mathcal{W}_p=1024$ pixels corresponding to each flexible instrument on the feature map. The U-Net model serves as the image encoder to extract image features, while a ResNet50~\cite{resnet} model act as the state decoder for regressing the state parameters. During training, we apply color jitter to 60\% of the images to augment the dataset. The robot state estimation model is trained for 100 epochs with a batch size of 24 on the NVIDIA RTX 3080 GPU. The training starts with an initial learning rate of $2\times10^{-4}$. Then, we incorporate a cosine schedule to anneal the learning rate at 60\% of the training epoch. 
{\bf{Depth:}} For monocular depth estimation, we utilize the ViT-base version from Depth Anything~\cite{yang2024depth} as our backbone. During training, the original image is cropped to $518\times518$ pixels as input. Our model is trained with the OneCycleLR strategy in which we select a maximum learning rate of $10^{-5}$ and train for 20 epochs with a batch size of 8 on eight NVIDIA RTX 2080Ti (12GB each). It takes around 24 hours to train our teacher network and approximately two days to the student network. 
The difference in training time is due to the fact that the student network was trained on unlabeled clinical data and was therefore exposed to significantly more data.
The following hyper-parameters for the training loss are chosen: $\lambda_{ssi}=1.0$, $\lambda_{re}=0.001$, $\alpha_{vn}=0.01$, $\lambda_{sup}=1.0$, and $\lambda_{ss}=1.0$.

\subsubsection{Datasets}
To train and evaluate each module of our proposed learning-based perception framework, we conducted ex-vivo trials to acquire endoscopic datasets using porcine GI tracts, including three colons and two stomachs.
We collected a total of six sets of video data, each recorded at 30 fps and with a resolution of $1020\times900$ pixels, yielding a total of 25000 frames.
For flexible instrument segmentation, we selected $4843$ frames for model training and testing. We manually labeled $971$ images using the LabelMe software for evaluation, while the remaining unlabeled data were utilized to train the annotation-efficient segmentation network. In robot state estimation, we used $9473$ frames for network training and $4000$ images for evaluating the performance of the flexible instrument state estimation model. For monocular depth estimation, we first generated $25000$ endoscopic images with corresponding depth maps from our proposed synthetic data generation framework. This simulated data was split into $17500$ for training, $5000$ for validation, and $2500$ frames for testing. Additionally, during the optimization of the student network, the entire set of 25000 unlabeled real endoscopic images was incorporated into the training process.

\subsection{Evaluation on 2D Continuum Robot Segmentation} 

\subsubsection{Competing Methods}
To evaluate the performance of our annotation-efficient segmentation framework, we conducted a comparative analysis against two baseline methods.
The first baseline, referred to as $\textbf{Syn}$, involved collecting synthetic data with ground truth annotations generated within the simulation environment. 
The second baseline, known as $\textbf{Copy-Paste}$~\cite{Copy-Paste}, utilized a data augmentation technique (described in Section~\ref{sec:segmentation}) applied to the synthetic data. 
This augmentation aimed to reduce the disparity between simulation and real-world scenarios. 
Both baseline methods were trained using synthetic data and subsequently tested on real clinical datasets.

\renewcommand{\arraystretch}{1.4}
\begin{table}[t]
\centering
\caption{Quantitative comparisons for 2D flexible instrument segmentation on ex-vivo data. $\uparrow$ indicates that higher values of the corresponding metric represent better performance.}
\label{tab:segmentation}
\begin{tabularx}{\columnwidth}{c|>{\centering\arraybackslash}p{0.6cm}>{\centering\arraybackslash}p{0.6cm} >{\centering\arraybackslash}p{1.1cm}|>{\centering\arraybackslash}p{0.6cm}>{\centering\arraybackslash}p{0.6cm}>{\centering\arraybackslash}p{1.0cm}} 
\specialrule{0.12em}{0pt}{2pt}
\multirow{2}{*}{Method} & \multicolumn{3}{c|}{Left} & \multicolumn{3}{c}{Right} \\ \cline{2-7} 
 & Dice~$\uparrow$ & IoU~$\uparrow$ & Precision~$\uparrow$ & Dice~$\uparrow$ & IoU~$\uparrow$ & Precision~$\uparrow$ \\ \midrule
\textbf{Syn} & 38.70 & 27.93 & 68.86 & 15.87 & 12.09 & 36.62 \\ \hline
\textbf{Copy-Paste} & 60.82 & 48.02 & 79.34 & 29.49 & 20.04 & 74.74 \\ \hline
\textbf{Ours} & \textbf{94.93} & \textbf{90.38} & \textbf{94.65} & \textbf{95.59} & \textbf{91.60} & \textbf{95.04} \\ 
\specialrule{0.12em}{2pt}{0pt}
\end{tabularx}
\end{table}

\begin{figure}[!ht]
\centering
\includegraphics[width = 0.95\hsize]{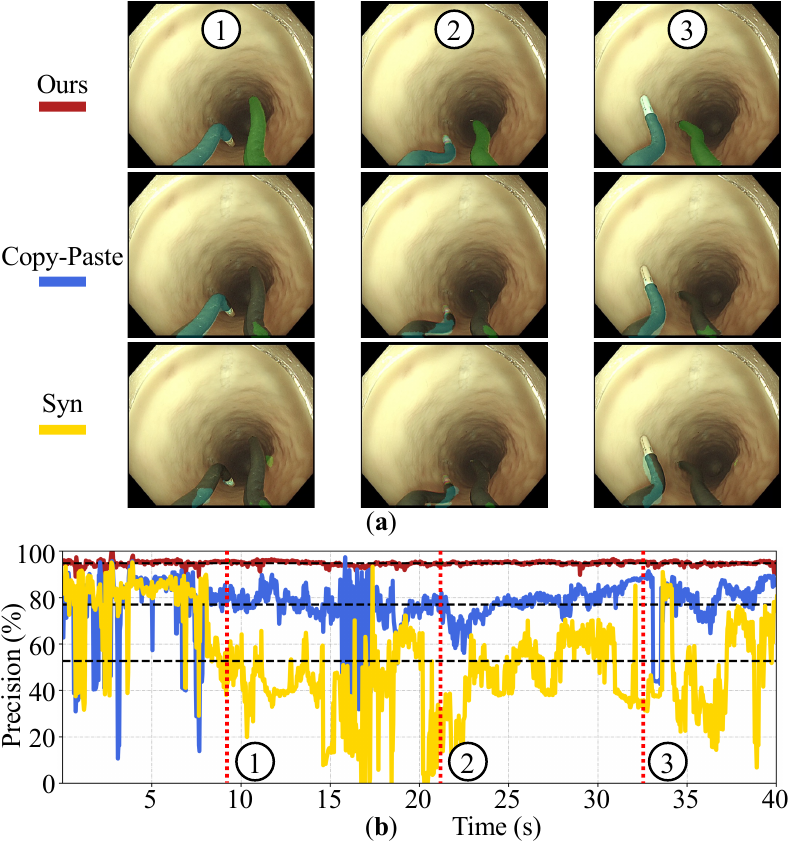}
\caption{Performance of flexible instrument segmentation tracking. (a) Examples of instrument segmentation visualization using different approaches during tracking. The semitransparent shadows indicate the predicted segmentations of the instruments. Our results align more accurately with the robotic configuration. (b) Precision comparison of the \textbf{Ours}, \textbf{Copy-Paste}~\cite{Copy-Paste}, and \textbf{Syn} segmentation methods against the ground truth. The numbered circles (1, 2, 3) indicate specific time points that align with the correspondingly numbered images in (a). Values represent the average precision across both left and right instrument segmentations.}
\label{fig:seg_result}
\vspace{-0.35cm}
\end{figure} 

\subsubsection{Metrics}
Following the guidelines from~\cite{segmetric}, we used three widely recognized evaluation metrics in computer vision to analyze model effectiveness: the Dice score, Intersection over Union (IoU), and Precision.
Additionally, we measured inference frames per second (fps) to evaluate model efficiency.

\renewcommand{\arraystretch}{1.4}
\begin{table}[t]
\centering
\caption{Quantitative comparison of our methods with the state-of-the-art methods on testing datasets.
}
\label{table:pose_result1}
\begin{tabular}{cc|c|c|c|c|c}
\specialrule{0.12em}{0pt}{2pt}
\multicolumn{2}{c|}{Metrics} & \textbf{KP} & \textbf{SKL} & \textbf{Quat}  & \textbf{Rot6D}   & \textbf{Ours}     \\ \specialrule{0.05em}{2pt}{2pt}
\multicolumn{1}{c|}{\multirow{2}{*}{$\alpha$}} & Mean $(^{\circ})\downarrow$     &   22.46  &  20.61   &  7.08   &  4.16       &     \textbf{2.60} \\
\multicolumn{1}{c|}{}                          & Med $(^{\circ})\downarrow$       &   24.45  &   22.70    &  3.09  & 2.43             &  \textbf{2.08} \\ \hline
\multicolumn{1}{c|}{\multirow{2}{*}{$\beta$}}  & Mean $(^{\circ})\downarrow$      &   25.82  &    19.50   &  7.11  & 5.54          &  \textbf{5.34} \\
\multicolumn{1}{c|}{}                          & Med $(^{\circ})\downarrow$       &   24.51  &  19.34   &  5.24  &  \textbf{3.83}               &  3.95 \\ \hline
\multicolumn{1}{c|}{\multirow{2}{*}{$\gamma$}} & Mean $(^{\circ})\downarrow$      &   29.89  &  17.70  &  8.31  &  7.09                &  \textbf{6.66} \\
\multicolumn{1}{c|}{}                          & Med $(^{\circ})\downarrow$       &  26.34  &  13.68   &  6.74   &  \textbf{5.44}           &  5.51 \\ \hline
\multicolumn{1}{c|}{\multirow{4}{*}{$\delta$}} & Mean $(^{\circ})\downarrow$      &  53.40  &  58.49   &   31.54   & \textbf{10.07}          & 11.09 \\
\multicolumn{1}{c|}{}                          & Med $(^{\circ})\downarrow$       &   35.18  &  47.18  &  28.58 &  \textbf{6.59}          &  7.38 \\
\multicolumn{1}{c|}{}                          & Acc$10^\circ\;(\%)\uparrow$  &   16.42   &  16.72   &   33.16   &  \textbf{75.01}            &   68.10\\
\multicolumn{1}{c|}{}                          & Acc$15^\circ\;(\%)\uparrow$ &   28.19   &  24.56   &  47.93   &  \textbf{86.54}           &  82.54 \\ \specialrule{0.12em}{2pt}{0pt}
\end{tabular}
\end{table}

\subsubsection{Results}
Table~\ref{tab:segmentation} presents the comparative results obtained from the two baseline methods alongside our proposed method. 
To ensure a fair and unbiased comparison, we matched the number of training data samples used for the baselines with those in our proposed method. 
Consistency in network architecture across all three methods was maintained by employing the U-Net. 
The $\textbf{Syn}$ baseline, which trained solely on synthetic data, demonstrated poor performance primarily due to the significant synthetic-to-real domain gap.
The substantial difference in data distribution between training and testing results in a notable drop in performance metrics.
In contrast, applying $\textbf{Copy-Paste}$ augmentation minimized the synthetic-to-real domain difference, allowing the model to better generalize its learned knowledge from synthetic data to real clinical scenarios, leading to enhanced performance on the test data. 
However, the performance of the model was still constrained by variations in factors such as robotic instrument configurations, lighting conditions, and texture differences between the synthetic and clinical endoscope images. 

To further assess the performance of our proposed flexible robot segmentation model across consecutive frames, we evaluated its ability to track flexible instrument segmentation over 40-second duration. As illustrated in Fig.~\ref{fig:seg_result}, our model maintained high and stable precision throughout the tracking period, making it suitable for the downstream robot state estimation. In contrast, the competing methods exhibited fluctuations and lower precision over time. Fig.~\ref{fig:seg_result}(a) visualizes typical examples of segmentation results, demonstrating that our model achieves superior performance across various robotic configurations. Additionally, we compared the inference speed between our model and the vision foundation model for supervision during training. 
Both models were tested on 1,000 images using an NVIDIA GeForce RTX 2080 Ti. 
The vision foundation model SAM with a ViT-Huge backbone achieved an inference speed of only 1.4\;fps. 
In contrast, our lightweight model demonstrated significantly improved efficiency, achieving an impressive inference speed of 52.1\;fps. 

\subsection{Evaluation on 3D Continuum Robot State Estimation}

\subsubsection{Competing Methods}
We compared our method against four SOTA baselines across three distinct categories.
Keypoint-based (\textbf{KP}):
We re-implemented the widely used keypoint method for robot state estimation as described in the work by DREAM~\cite{DREAM}. 
We defined six predetermined keypoints on the flexible robot body and employed a SOTA image-based keypoint localization method~\cite{track_anything} to accurately locate these keypoints in the images. 
Using the extracted keypoints, we determined the shape of the flexible robot and computed the corresponding state parameters.
Skeleton-based (\textbf{SKL}): 
We implemented a method derived from the skeleton extraction approach outlined in~\cite{skeleton_1}. 
We first extracted the skeleton of the flexible instruments from its binary mask using a fast skeletonization algorithm.
We then applied a robust fitting technique, specifically the Bezier curve, to accurately compute the robot state parameters from the skeleton representation.
Regression-based (\textbf{Quat} and \textbf{Rot6D}):
In this category, we conducted a comparative analysis between our proposed method and two alternative rotation representation methods: quaternion and rotation six-dimensional (6D). 
For the quaternion representation (\textbf{Quat}), we followed the methodology outlined in~\cite{SimPS} to represent each state parameter using quaternions.
Similarly, for the rotation 6D approach (\textbf{Rot6D}), we employed a method akin to that described in~\cite{rot6d,CoRL}, utilizing a 6D rotation encoding scheme.
To ensure a fair and unbiased comparison, we maintained a consistent network architecture across all methods, adjusting only the state representation head to accommodate the respective rotation representations.

\subsubsection{Metrics}
For quantitative evaluation, we followed the previous work of~\cite{fishermatch} to present the average and median angular errors for each predicted state parameter. 
Additionally, we reported prediction accuracy in relation to $5^{\circ}$ and $10^{\circ}$. 
This metric quantifies the proportion of predictions with errors smaller than these values.
For qualitative comparison, we visualized the estimated 3D robot state of the flexible instruments within 2D images using our proposed simulator.
We controlled the motion of the simulated flexible instruments according to the estimated robot states. 
Then, the instruments are projected onto 2D coordinates system and rendered into the endoscopic images using the camera extrinsic and intrinsic matrices.

\subsubsection{Results}

\begin{figure}[t!]
\centering
\includegraphics[width = 0.95\hsize]{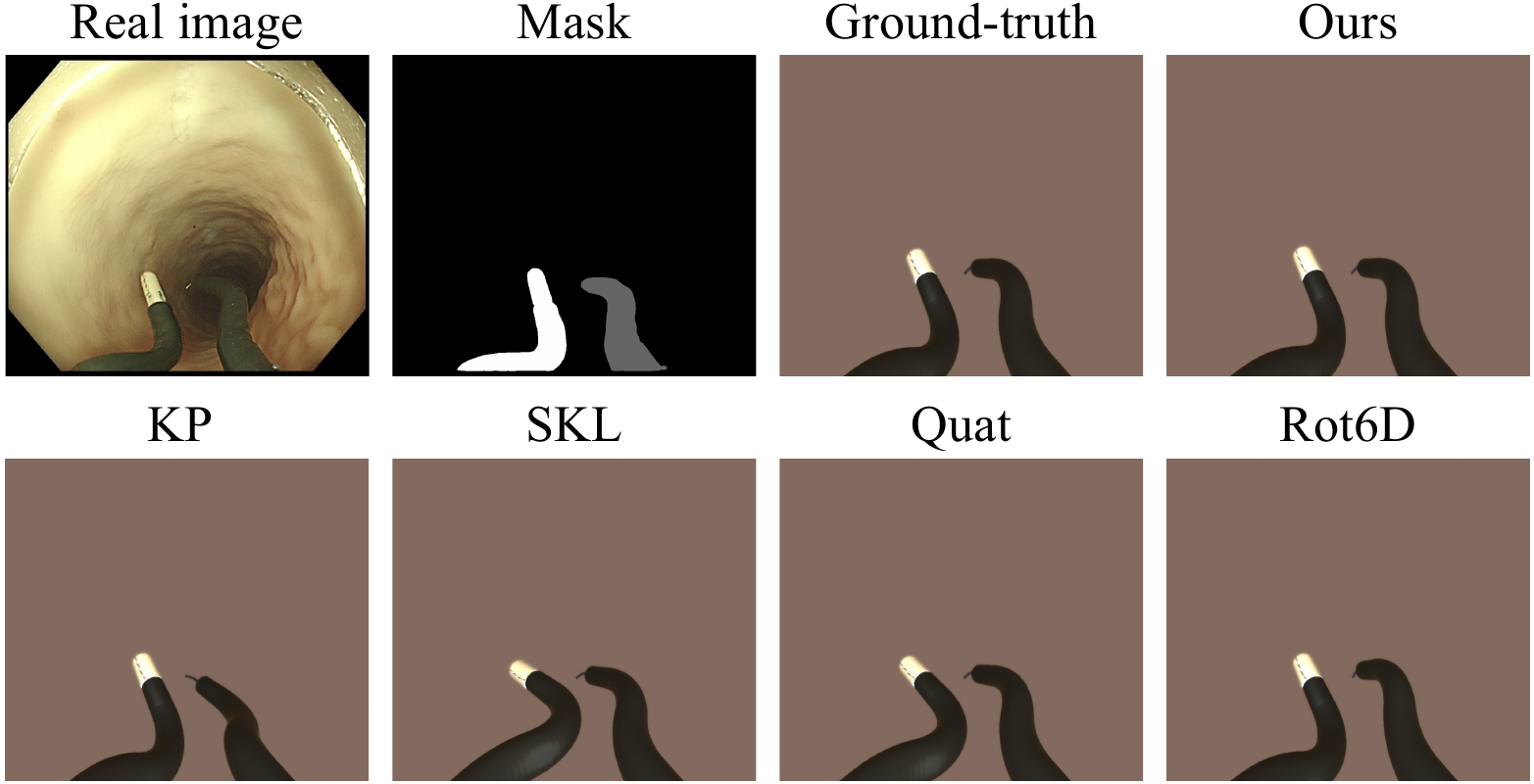}
\caption{Qualitative comparison of our continuum robot state estimation model with the state-of-the-art methods:
keypoint-based \textbf{KP}~\cite{DREAM}, skeleton-based \textbf{SKL}~\cite{skeleton_1}, quaternion pose representation \textbf{Quat}~\cite{SimPS}, and rotation 6D pose representation \textbf{Rot6D}~\cite{CoRL}.
First, the flexible instruments were manipulated using the estimated state parameters within the proposed simulator. Then, the simulated instruments were projected onto 2D coordinates system to render synthetic images, utilizing the camera extrinsic and intrinsic matrices.}
\label{fig:pose_qualitative}
\vspace{-0.35cm}
\end{figure} 

Table~\ref{table:pose_result1} and Fig.~\ref{fig:pose_qualitative} present quantitative and qualitative comparisons between our method and the four SOTA approaches on the testing datasets. 
Among these methods, the keypoint-based approach (\textbf{KP}) demonstrates poor performance, primarily due to instability in detecting and tracking sparse keypoints on textureless continuum robots. 
Consequently, the robot state calculation becomes unreliable.
In contrast, the skeleton-based method (\textbf{SKL}) achieved improved performance by representing the robotic instrument as a complete skeleton. 
This representation captures more geometric information than discrete keypoints. 
However, the accuracy of robot state estimation was still limited by challenges in skeleton extraction, particularly due to the high DoF of flexible instruments.
On the other hand, the regression-based methods (\textbf{Quat} and \textbf{Rot6D}), leveraging the strong learning capacity of deep neural networks, significantly outperformed both \textbf{KP} and \textbf{SKL}. 
The image encoder effectively captured intricate patterns and implicit information from images, modeling the complex relationship between image features and robot state parameters.
Nevertheless, the inherent discontinuity in quaternion representation posed challenges for neural networks in learning from loss functions. 
While \textbf{Rot6D} addresses this issue and achieves comparable performance to our method, our approach goes a step further. 
We obtain the rotation matrix and its uncertainty from a probabilistic distribution, which is potentially valuable for robot manipulation tasks.

\begin{figure}[t]
\centering
\includegraphics[width = 0.95\hsize]{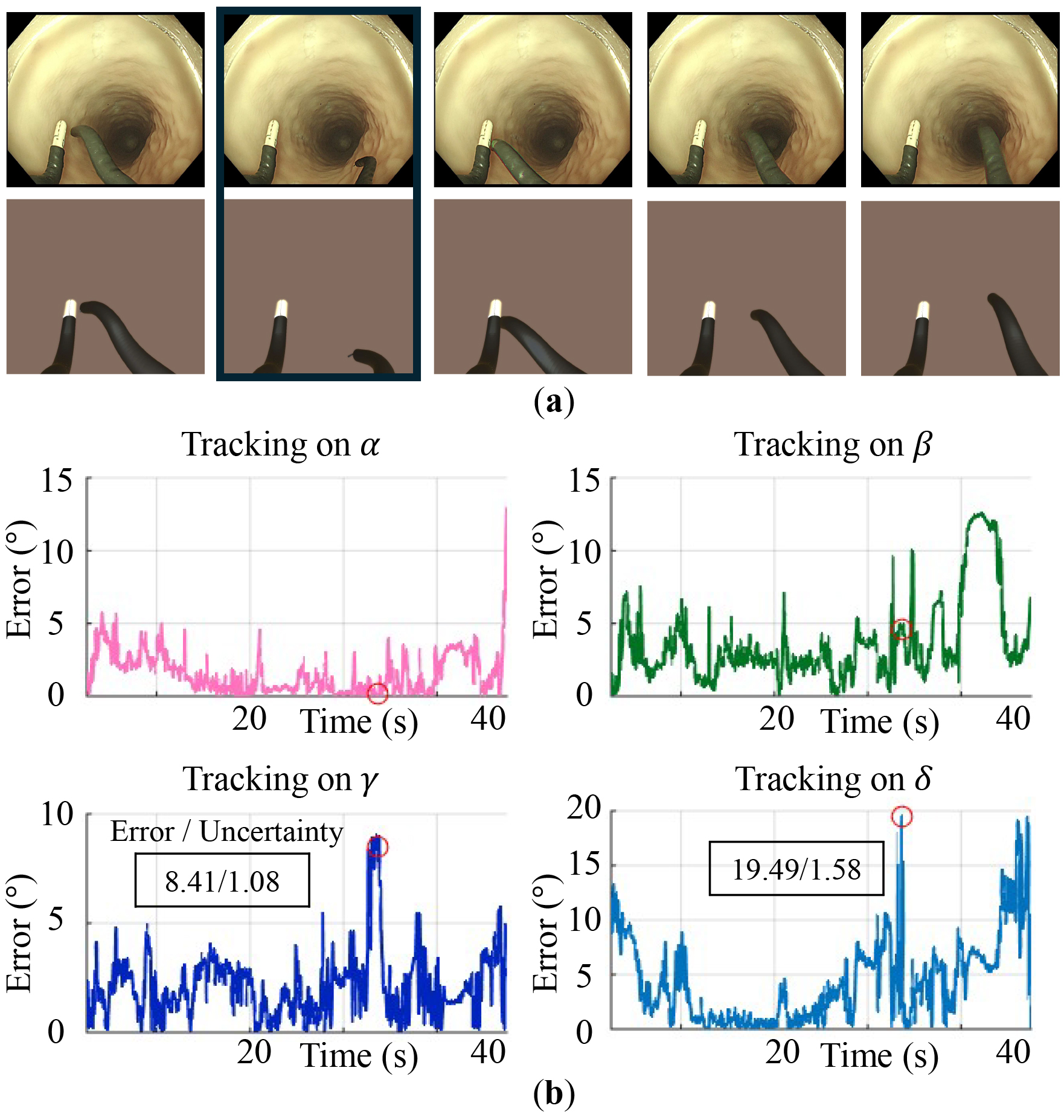}
\caption{Performance of continuum robot state tracking. (a) Comparison examples between real and rendered endoscopic images. (b) Variations in tracking error across all estimated robot states over time. The time corresponding to the second frame in (a) with high error is highlighted with a red circle, and its error and uncertainty are presented.}
\label{fig:uncertain_left}
\vspace{-0.35cm}
\end{figure} 

\begin{figure*}[t]
\centering
\includegraphics[width = 0.98\hsize]{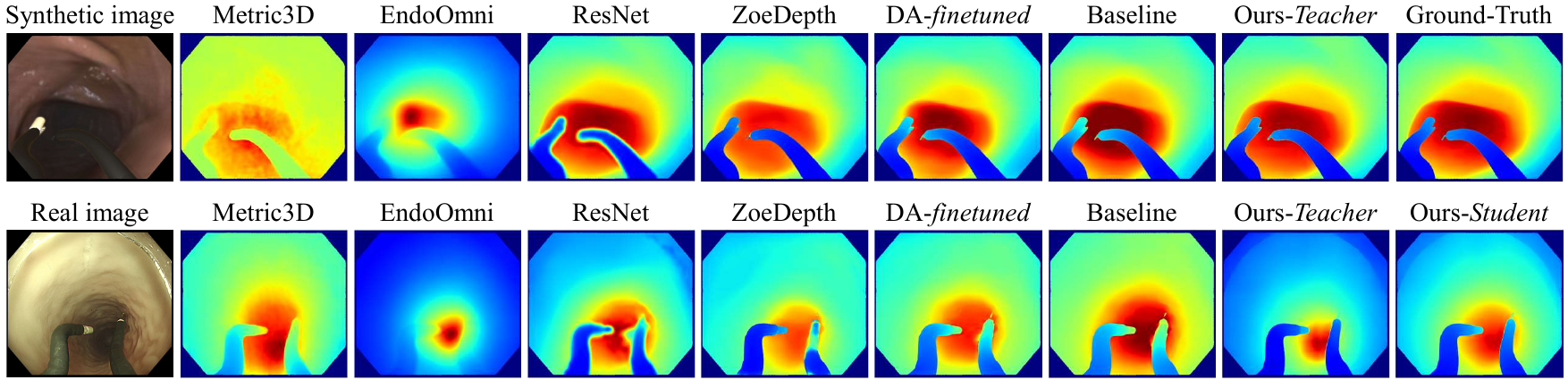}
\caption{Qualitative comparison of monocular depth estimation on synthetic and real endoscopic data. The depth estimation results of our model are compared with foundation models Metric3D~\cite{yin2023metric3d} and EndoOmni~\cite{tian2024endoomni}, finetuned Depth Anything (DA-\textit{finetuned})~\cite{yang2024depth}, ZoeDepth~\cite{bhat2023zoedepth} and ResNet~\cite{resnet}. Each row shows flexible robots in the same configuration. Our method captures depth with fine details, where \textit{\color{red}red} indicates greater distance from the endoscope and \textit{\color{blue}blue} indicates proximity.}
\label{fig:depth_comparison}
\vspace{-0.2cm}
\end{figure*}

Following the methodology in~\cite{fishermatch}, we utilized entropy as a measure of uncertainty in our flexible robot state estimation model.
To improve the visual representation of our uncertainty measurements, we normalized the originally negative values by adding 6 and then dividing by 1.5. 
This transformation establishes a threshold of 1.0 for data filtering.
Fig.~\ref{fig:uncertain_left} illustrates the performance of our proposed model across consecutive frames during a controlled movement of the right instrument over approximately 40 seconds.
During this tracking process, we calculated the error between our predictions and the ground truth. The error variations throughout the tracking period are depicted in Fig.~\ref{fig:uncertain_left}(b). We also selected five key frames for comparison between the rendered and real endoscopic images, as shown in Fig.~\ref{fig:uncertain_left}(a).
The errors and uncertainty values of the second key frame are presented.
Our observations indicate that predictions associated with lower uncertainty generally correspond to lower error rates. Conversely, predictions with higher uncertainty tend to produce a higher frequency of significant errors. This finding underscores the effectiveness of our uncertainty estimation in reflecting the quality of robot state estimation.
Additionally, during the last 12 seconds of tracking, the left instrument kept static, and we noted that the robot state estimation maintained stability, with uncertainty levels hovering around 0.4. This stability reinforces our model's reliability in dynamic and static scenarios alike.

\subsection{Evaluation on 3D Scene Depth Estimation} 

\subsubsection{Competing Methods}
We compared our proposed monocular depth estimation model with five baseline methods. ResNet~\cite{resnet} demonstrates strong performance when trained with ground-truth depth labels, as it excels at extracting multi-scale contextual information and localizing features accurately through its symmetric encoder-decoder design. We trained ResNet on our synthetic continuum robot surgical data using supervised depth loss for 20 epochs. We also evaluated ZoeDepth~\cite{bhat2023zoedepth}, a unified multi-domain depth estimation model designed for cross-domain generalization. ZoeDepth was trained on our simulated endoscopic dataset using scale-invariant logarithmic loss for 20 epochs. 
Depth Anything~\cite{yang2024depth} was fine-tuned on our synthetic data for 5 epochs to assess its adaptability to surgical scenes. Additionally, we evaluated two zero-shot models: Metric3D~\cite{yin2023metric3d}, a single-view monocular depth estimator trained on large-scale mixed datasets, and EndoOmni~\cite{tian2024endoomni}, a foundation model specifically designed for cross-domain depth estimation in endoscopy. Both models were applied directly to our robotic endoluminal data without fine-tuning to assess their generalization capabilities in surgical scenarios.
Following this, we introduced our baseline model, which fine-tunes a pre-trained DINOv2 encoder~\cite{oquab2023dinov2} with a DPT-based decoder~\cite{ranftl2021vision} using supervised depth loss on the synthetic endoscopic data labels, excluding our supervised PPR loss $\mathcal{L}_{sup}$. We then present the Ours-$Teacher$ network, which integrates our baseline with the $\text{PRMod}$ depth improvement module and is trained on the proposed synthetic data with both supervised depth and PPR losses. Finally, the Ours-$Student$ network has the same structure as Ours-$Teacher$ but is trained on both synthetic and real clinical data using supervised loss $\mathcal{L}_{sup}$ and self-supervised PPR loss $\mathcal{L}_{ss}$, respectively.

\subsubsection{Metrics}

\setlength{\tabcolsep}{4pt}
\renewcommand{\arraystretch}{1.6}
\begin{table}[t!]
\centering
\caption{Depth Evaluation Metrics}
\label{table_depth_metrics} 
\begin{tabular}{c c}
\specialrule{0.12em}{0pt}{0pt}
Metrics & Definition \\
\hline
Abs Rel &  $\frac{1}{|\bm{D}|} \sum_{d\in\bm{D}}{|d^* - d|/d^*}$\\
Sq Rel & $\frac{1}{|\bm{D}|} \sum_{d\in\bm{D}}{|d^*-d|^2/d^*}$ \\
RMSE & $\sqrt{\frac{1}{|\bm{\mathcal{D}}|} \sum_{d\in\bm{\mathcal{D}}}{|d^*-d|^2}}$\\
$\text{RMSE}_{log}$ & $\sqrt{\frac{1}{|\bm{D}|} \sum_{d\in\bm{D}}{|\log d^*-\log d|^2}}$\\
$\eta$ & $\frac{1}{|\bm{D}|}\left\{ d \in \bm{D} |\max(\frac{d^*}{d},\frac{d}{d^*} < \upsilon)|\right\} \times 100\%$\\
\specialrule{0.12em}{2pt}{2pt}
\end{tabular}
\vspace{-0.1cm}
\end{table}

Table~\ref{table_depth_metrics} outlines the depth evaluation metrics employed in our experiments~\cite{godard2019digging}. Here, $d$ represents the predicted depth value, $d^*$ the corresponding ground truth, $\bm{D}$ denotes the set of predicted depth values, and $\upsilon \in \{1.25^1, 1.25^2, 1.25^3\}$.

\setlength{\tabcolsep}{4pt}
\renewcommand{\arraystretch}{1.0}
\begin{table*}[t]
\caption{Quantitative comparisons of monocular depth estimation on synthetic robotic endoluminal test data. GT, ground truth; SSL, self-supervised learning. The best results are in bold. All values in \textit{mm}.}
\label{table: quantitative_depth}
\centering
\begin{tabularx}{\textwidth}{c|c|c|>{\centering\arraybackslash}X>{\centering\arraybackslash}X>{\centering\arraybackslash}X>{\centering\arraybackslash}X|>{\centering\arraybackslash}X>{\centering\arraybackslash}X>{\centering\arraybackslash}X} 
\toprule
\multirow{2}{*}{\text{Method}} & \multirow{2}{*}{\text{Finetuning}} & \multirow{2}{*}{\text{Supervision}}& \multicolumn{4}{c|}{$\text{Error} \downarrow$}       & \multicolumn{3}{c}{$\text{Accuracy} \uparrow$} \\ \cline{4-10} 
                        &            &            & \text{Abs Rel} & \text{Sq Rel} & \text{RMSE}  & $\text{RMSE}_{log}$ & $\eta\!<\!1.25^1$ & $\eta\!<\!1.25^2$ & $\eta\!<\!1.25^3$ \\ \midrule
Metric3D~\cite{yin2023metric3d} & \XSolidBrush & $-$ & 0.239  & 3.671 & 12.543 & 0.298 & 0.494 & 0.847 & 0.986 \\ 
EndoOmni~\cite{tian2024endoomni} & \XSolidBrush & $-$ & 0.273  & 4.048 & 11.370 & 0.354 & 0.474 & 0.769 & 0.92 \\
ResNet~\cite{resnet} & \Checkmark & GT & 0.169  & 1.490 & 7.703 & 0.255 & 0.712 & 0.917 & 0.965 \\ 
ZoeDepth~\cite{bhat2023zoedepth} & \Checkmark & GT & 0.114 & 0.857 & 6.210 & 0.174 & 0.867 & 0.953 & 0.999 \\
Depth Anything~\cite{yang2024depth} & \Checkmark & GT & 0.113  & 0.806 & 5.094 & 0.160 & 0.858 & 0.952 & 0.995 \\ \midrule
Baseline & \Checkmark & GT  & 0.091  & 0.653 & 4.655 & 0.146 & 0.858 & 0.977 & 0.997 \\ 
Ours & \Checkmark & GT \& SSL &  \textbf{0.069} & \textbf{0.365} & \textbf{4.020} & \textbf{0.099} & \textbf{0.950} & \textbf{0.999} & \textbf{1.000} \\ 
\bottomrule
\end{tabularx}
\vspace{-0.2cm}
\end{table*}

\subsubsection{Results}
Table~\ref{table: quantitative_depth} presents the quantitative depth comparison results on our synthetic endoluminal endoscopic data. Our proposed model achieves the best performance across all depth evaluation metrics, with an RMSE of approximately 4.020 \textit{mm}. This indicates high accuracy in estimating the depth of tissue surfaces and flexible instruments. Furthermore, the enhanced depth evaluation results suggests that our proposed loss functions and synthetic-to-real transfer learning are beneficial for depth estimation in endoluminal and endoscopic surgical scenarios. The quantitative results for the depth estimation foundation models indicate that training on large-scale, domain-specific data significantly boosts performance. Additionally, we selected one image from the simulated surgical data and one frame from the real robotic endoluminal surgery for qualitative depth comparison. As shown in Fig.~\ref{fig:depth_comparison}, our method generates high-quality depth maps with fine-grained details, effectively capturing features such as the tips of flexible instruments and the geometry of tissue surfaces. 

\begin{figure}[t]
    \centering
    \includegraphics[width = 0.95\hsize]{"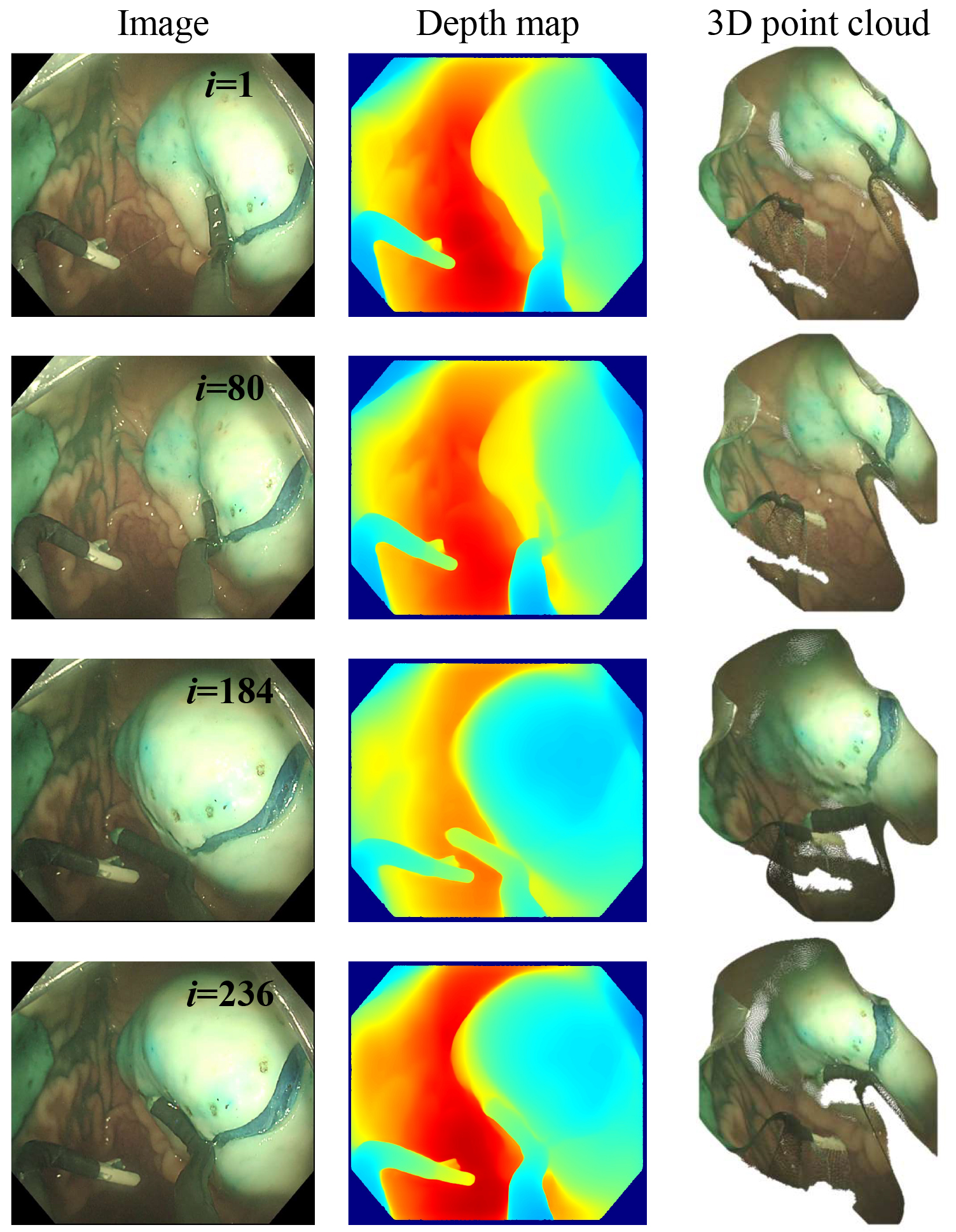"}
    \caption{Reconstructed 3D point cloud from the monocular depth estimation of real continuum robotic endoluminal surgical data. Variable $i$ denotes the frame index in the video sequence. More results of monocular depth estimation can be found in Video 3.}
    \label{fig:3d_dis}
    \vspace{-0.35cm}
\end{figure}

Moreover, as illustrated in Fig.~\ref{fig:3d_dis}, we converted the depth estimates from our monocular depth estimation model into 3D point clouds to analyze the configuration changes of flexible robots and their relationship with tissue surfaces. During these consecutive frames, the flexible instrument was controlled to push against an elevated cushion on the colon phantom. The smooth and geometrically consistent 3D structures clearly reveal the 3D shape deformation of the right flexible instrument and the tissue surface. Furthermore, the spatial relationship between the instrument and tissue is readily discernible from the 3D point cloud. 
Without explicit 3D structure, understanding the spatial relationships between objects is challenging. For instance, viewing only the image at frame $i=1$ does not reveal whether the right instrument is in contact with the colon surface. However, the 3D point cloud clearly shows that the instrument push deforms the elevated tissue region. Observing the sequence from frames $i=1$ to $i=236$ reveals the complete interaction: the instrument initially pushes the tissue downward, is then retracted, and subsequently makes contact again with reduced tissue deformation. This ability to infer instrument-tissue interactions from monocular depth estimation is particularly valuable for autonomous surgery in constrained spaces using continuum robotic systems.

\subsubsection{Ablation Study}
To better understand the contributions of the depth improvement module and the supervision PPR loss to the overall depth estimation performance, we conducted an ablation study on \textbf{PRMod} and $\mathcal{L}_{sup}$. As shown in Table~\ref{table: quantitative_depth} and Fig.~\ref{fig:depth_comparison}, Ours-$Teacher$ demonstrates better performance than our baseline model. Afterward,  we proposed two distinct enhancements for the baseline: the depth improvement module \textbf{PRMod} and the loss $\mathcal{L}_{sup}$. Table~\ref{table: ablation_ppr} illustrates that both $\mathcal{L}_{sup}$ and \textbf{PRMod} significantly enhance the capabilities of Ours-$Teacher$ compared to baseline.

\setlength{\tabcolsep}{4pt}
\renewcommand{\arraystretch}{1.0}
\begin{table}[t]
\caption{Ablation study on $\mathcal{L}_{sup}$ and \textbf{PRMod}.}
\label{table: ablation_ppr}
\centering
\begin{tabular}{cc|cccc|c}
\specialrule{0.12em}{0pt}{2pt}
\multirow{2}{*}{$\mathcal{L}_{sup}$} & \multirow{2}{*}{\textbf{PRMod}} & \multicolumn{4}{c|}{$\text{Error} \downarrow$}       & {$\text{Accuracy} \uparrow$} \\ \cline{3-7} 
                        &                        & \text{Abs Rel} & \text{Sq Rel} & \text{RMSE}  & $\text{RMSE}_{log}$ & $\eta\!<\!1.25^1$  \\ \specialrule{0.02em}{2pt}{2pt}
\XSolidBrush & \XSolidBrush  & 0.096  & 0.718 &  4.864 & 0.150 & 0.842 \\ 
\Checkmark & \XSolidBrush & 0.088 & 0.610 & 4.576 & 0.145 & 0.870 \\
\XSolidBrush & \Checkmark  &  0.093 & 0.664 & 4.728 & 0.146 & 0.858 \\ 
\Checkmark & \Checkmark   &  \bf{0.072} & \bf{0.395} & \bf{4.286}  &  \bf{0.102} & \bf{0.945} \\ \specialrule{0.12em}{2pt}{0pt}
\end{tabular}
\vspace{-0.1cm}
\end{table}

We further analyzed the impact of the image features $\bm{\mathcal{F}}_{\text{img}}$ and PPR features $\bm{\mathcal{F}}_{\text{PPR}}$ on depth estimation. As listed in Table~\ref{table: ablation_features}, we found that adjusting the initial depth features extracted from endoscopic images using DINOv2, along with RGB features and PPR representation, boosts the effectiveness of the depth improvement module.

\setlength{\tabcolsep}{4pt}
\renewcommand{\arraystretch}{1.0}
\begin{table}[t]
\caption{Ablation study on image features $\bm{\mathcal{F}}_{\text{img}}$ and PPR features $\bm{\mathcal{F}}_{\text{PPR}}$.}
\label{table: ablation_features}
\centering
\begin{tabular}{cc|cccc|c}
\specialrule{0.12em}{0pt}{2pt}
\multirow{2}{*}{$\bm{\mathcal{F}}_{\text{img}}$} & \multirow{2}{*}{$\bm{\mathcal{F}}_{\text{PPR}}$} & \multicolumn{4}{c|}{$\text{Error} \downarrow$}       & {$\text{Accuracy} \uparrow$} \\ \cline{3-7} 
                        &                        & \text{Abs Rel} & \text{Sq Rel} & \text{RMSE}  & $\text{RMSE}_{log}$ & $\eta\!<\!1.25^1$  \\ \specialrule{0.02em}{2pt}{2pt}
\Checkmark & \XSolidBrush  & 0.074  & 0.439 & 4.486 & 0.107 & 0.936 \\ 
\XSolidBrush & \Checkmark  & 0.073 & 0.399 & 4.323 & 0.105 & 0.943 \\
\Checkmark & \Checkmark  & \bf{0.072}  & \bf{0.395} & \bf{4.286} & \bf{0.102} & \bf{0.944} \\ \specialrule{0.12em}{2pt}{0pt}
\end{tabular}
\vspace{-0.1cm}
\end{table}

Finally, Fig.~\ref{fig:depth_comparison} shows that Ours-$Student$ outperforms Ours-$Teacher$, particularly on real clinical data. This improvement indicates that while Ours-$Teacher$ was trained solely on synthetic data and struggles to generalize well to real data, the teacher-student transfer learning approach enhance the quality of monocular depth estimation in real continuum robotic endoluminal surgical scenes.

\subsection{Runtime}
To systematically evaluate the runtime performance of the proposed monocular image-based perception framework, all modules were executed on Alienware desktop with an AMD Ryzen 9 7900 CPU and an NVIDIA RTX 4090 (24GB). Performance was assessed using a standard endoscopic video ($1020 \times 900$ resolution, 5418 frames, approx. 3 minutes). For 2D segmentation and 3D robot state estimation, input frames were resized to $224 \times 224$. For 3D depth estimation, the input size was set to $518 \times 518$. As shown in Table~\ref{tab:runtime}, the 2D segmentation and 3D state estimation modules are executed sequentially, while the 3D depth estimation branch runs in parallel. Consequently, the overall system latency is determined by the depth estimation bottleneck, resulting in an average processing time of 47.15 ms per frame. This corresponds to a frame rate of $\sim$21 fps, meeting the requirements for real-time endoscopic feedback.

\begin{table}[t]
\centering
\caption{Runtime performance of monocular image-based 2D and 3D perception framework.}
\label{tab:runtime}
\begin{tabular}{c|c}
\toprule
Modules & Execution time (ms) \\ 
\midrule
2D segmentation & 5.50 \\
3D state estimation & 20.72 \\
\textit{Subtotal (sequential)} & \textit{26.22} \\
\midrule
3D depth estimation (parallel) & 47.15 \\ 
\midrule
\textbf{Total system latency} & \textbf{47.15} \\
\bottomrule
\end{tabular}
\vspace{-0.1cm}
\end{table}
 
\section{Real-world Evaluation of Image-based Perception on Continuum Robotic System}
\label{sec:application}

In this section, we propose three automatic cognitive assistance functions to evaluate our image-based perception modules on the continuum robotic system for endoluminal surgery, thereby addressing challenges in remote control and enhancing surgeons' clinical decision-making. These cognitive assistance functions are incorporated into a novel continuum robotic system, which comprises disposable flexible surgical instruments, a robotic position cart, an Olympus endoscope imaging system, and a compact control console.

The flexible robotic instruments, which feature five degrees of freedom, are specifically designed to grip, cut, and completely remove tumor tissue with exceptional precision within natural orifices. Notably, these instruments are fully compatible with standard endoscopes commonly used in hospitals. Additionally, the endoscope holder in the positioning cart offers surgeons significant flexibility, allowing for optimal positioning of both the endoscope and the robotic instruments. During operations, the surgeon can easily reposition and adjust the endoscope thanks to the holder's switchable locking feature, which ensures stability while facilitating movement. 
Furthermore, the flexible instruments are controlled by the surgeon seated at the control console, who utilizes two pen-like controllers to manipulate the robots' movements intuitively. 

The cognitive assistance functions consist of: (1) remote-control calibration of the flexible robots to ensure precise operation, (2) 3D robot status monitoring of the instruments to provide real-time feedback on their position, and (3) instrument-tissue distance inference between instruments and tissue to enhance spatial awareness during surgery. These perception-based functions leverage advanced learning-based techniques, including 2D segmentation of flexible robotic instruments, 3D image-based robot state estimation, and 3D monocular endoscopic scene depth estimation. By employing these automatic functions, the robotic instruments can be manipulated with high precision in the confined endoluminal surgical environment, ultimately leading to improved surgical outcomes.

\subsection{Flexible Robot Remote-Control Calibration}

Flexible robot remote-control calibration aims to automatically establish the mapping between the robotic instruments and the controller when the surgical field is adjusted by rotating the endoscope. Typically, when the surgeon moves the controller (either up and down or left and right), the instruments respond with corresponding actions in the endoscopic view. However, to target specific areas outside the surgeon's direct line of sight, the endoscope must be rotated to adjust the field of view (FoV). As the instruments are delivered through the tool channels of the steerable endoscope, they rotate accordingly, which alters the relationship between the controller and the instruments. In such cases, the flexible instruments may not respond as expected due to this altered mapping. Consequently, if calibration is not performed, the surgeon may need to spend significantly additional time adapting to the abnormal robot-controller mapping. To address this issue, we propose using the robot state estimation model to calibrate the anomalous mapping. Specifically, the roll information obtained from the robot state estimation is integrated into the continuum robotic system, complementing the endoscope's rotation and ensuring accurate calibration.

To evaluate the importance and efficiency of the flexible robot remote-control calibration, we conduct user studies with two groups: one consisting of professionals experienced in manipulating flexible robotic instruments, and the other comprising inexperienced users who have received instructions on controlling the robot. Participants are tasked with manipulating the flexible instruments to follow a curved trajectory, with trials conducted both with and without calibrating the continuum robotic system. The trajectory marked on the colon surface involves 10 target points, with approximately equal distances. This curved trajectory-following task serves as an exercise to assess surgeon performance in controlling the flexible instruments during the endoluminal surgery. We will record the manipulation time for each trail to facilitate a comprehensive evaluation of outcomes. When the mapping between the instruments and controller is anomalous and the calibration is not performed, the professional group takes an average of 233 $s$ to complete the task, while the inexperienced users require approximately 254 $s$. This indicates that professionals are adept at handling the discrepancies in control but still face significant delays. Following this, we perform calibration informed by our robot state estimation results. After calibration, the manipulation time for professional users decreases dramatically to 66 $s$, while the inexperienced users reduce their time to about 72 $s$. The results clearly demonstrate the effectiveness of our calibration module. The substantial reduction in manipulation time—over 71\% for professionals and around 72\% for inexperienced users—highlights its impact on operational efficiency. This improvement suggests that calibration significantly reduces the cognitive load on users, allowing them to perform tasks intuitively. Overall, these findings underscore the critical role of calibration in optimizing the performance of the continuum robotic system during endoluminal surgery.

\subsection{Flexible Robotic Instrument 3D Shape Monitoring}

\begin{figure}[t]
\centering
\includegraphics[width = 0.95\hsize]{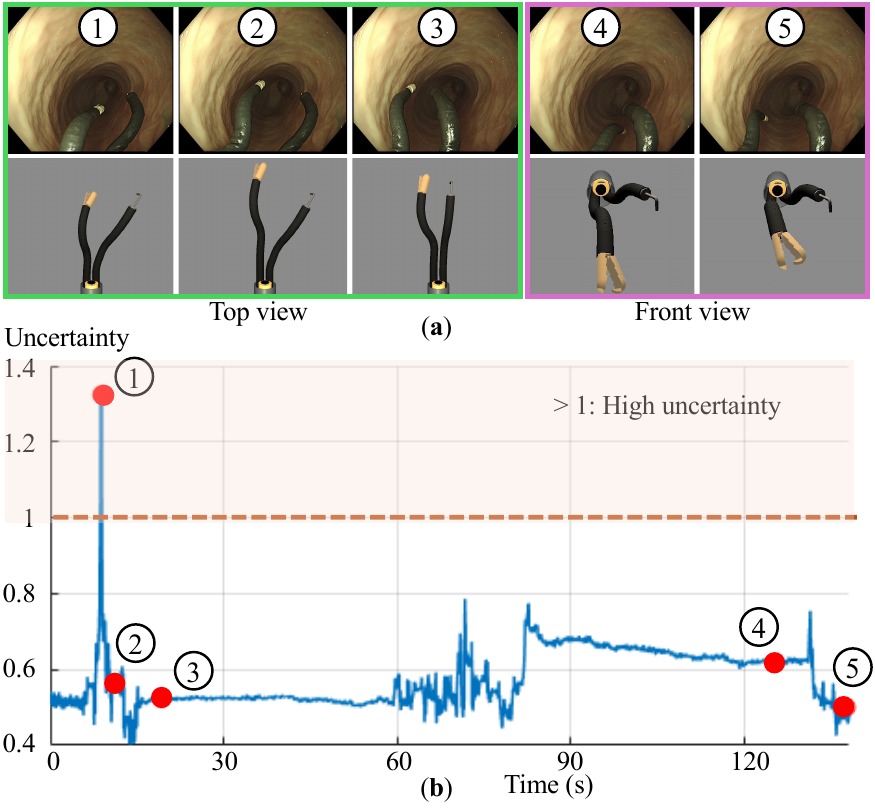}
\caption{Flexible instruments 3D shape monitoring. (a) Examples of endoscopic views alongside the corresponding robotic shapes visualized in the software. The 3D shapes of two instruments are shown from both top and front views. (b) Uncertainty in robot state estimation over time, with fives shapes from (a) marked. Areas with uncertainty values greater than 1 are overlaid with a semi-transparent orange color.}
\label{fig:shape_display}
\vspace{-0.3cm}
\end{figure}

Monitoring the 3D shape of flexible instruments requires a comprehensive understanding of their robotic states, denoted as $(\alpha, \beta, \gamma, \delta)$, as illustrated in Fig.~\ref{fig:simulator}(b). To achieve this, we developed 3D visualization software based on our physically-realistic simulator. In this software, the predicted flexible instruments states are displayed in 3D space. Additionally, the uncertainty from the proposed robot state estimation module is presented to reflect the quality of the predictions, particularly regarding the uncertainty of the roll. 
As shown in Fig.~\ref{fig:shape_display}, we manipulate the flexible instruments to control their contact with the colon surface, simulating the surgeon operations. Initially, when the Shoulder $\widehat{\textbf{OS}}$ of the left instrument [marked with \textcircled{\small{1}} in Fig.~\ref{fig:shape_display}(a)] moves out of the view, the model indicates a higher uncertainty in state estimation; however, the rendered 3D shape still closely resembles the actual view. When we insert the left instrument [marked with \textcircled{\small{2}}], the entire structure become visible, leading to a decrease in uncertainty. Subsequently, as we maneuver the right instrument toward the center [from \textcircled{\small{2}} to \textcircled{\small{3}}], a similar trend in estimation accuracy is observed from the software. For the remaining frames[marked as \textcircled{\small{4}}, \textcircled{\small{5}}], we display the 3D shapes of the instruments from the front view. 
As shown in Fig.~\ref{fig:shape_display}(b), throughout the manipulation process, the uncertainty remains low except when the part of the left instrument is out of view, indicating that the estimated states are stable and accurate. This is further confirmed by the comparison between the actual and rendering instruments shapes in 2D image.  
Overall, the results demonstrate that our robot state estimation effectively provides accurate representations of the instruments' states, while the visualization software offers clear and diverse 3D views for monitoring theirs shapes. In instances of abnormal situations, our model can still generate a rough prediction, albeit with higher uncertainty values. 

\begin{figure}[t]
\centering
\includegraphics[width = 0.95\hsize]{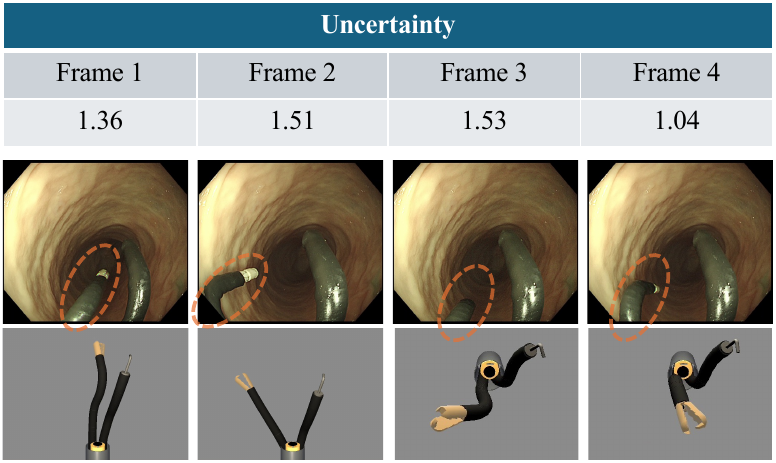}
\caption{Failure cases in robot state estimation. Instruments marked with dashed ovals illustrate how partial visibility results in incorrect state estimation and increased uncertainty. 
}
\label{fig:shape_failure}
\vspace{-0.35cm}
\end{figure} 

Fig.~\ref{fig:shape_failure} presents several failure cases that highlight the limitations of the robot state estimation model. These cases were selected during the execution of flexible instruments in an ex-vivo colon setting. In Frame 1, part of the robot is visible, similar to what is shown in Fig.~\ref{fig:shape_display}, resulting in a rough estimation characterized by high uncertainty. As the left instrument is manipulated to touch the colon surface on the left side, a large portion of it goes out of the view. Consequently, incorrect robotic states are predicted, leading to an uncertainty value of 1.51. Similarly, in Frame 3, the left instrument appears to blend with the background and a significant part of the robot disappears. This results in an even higher uncertainty of 1.53. An interesting observation occurs between Frame 3 and 4: as the instrument is rotated back into the view, a small part remains out of sight. Despite this partial visibility, the estimation model predicts relatively accurate robotic states with an uncertainty of 1.04. This suggests that the model lack confidence in the current estimation due to the instrument's incomplete visibility, highlighting the instability of the estimation in such scenarios.
In conclusion, these failure cases demonstrate that the robot state estimation model's accuracy is significantly affected by the visibility of the instruments. Situations where parts of the robot are obscured lead to increased uncertainty and less reliable predictions, underscoring the need for enhanced algorithms to improve stability and confidence in state estimations.

\begin{figure*}[t]
    \centering
    \includegraphics[width = 0.95\hsize]{"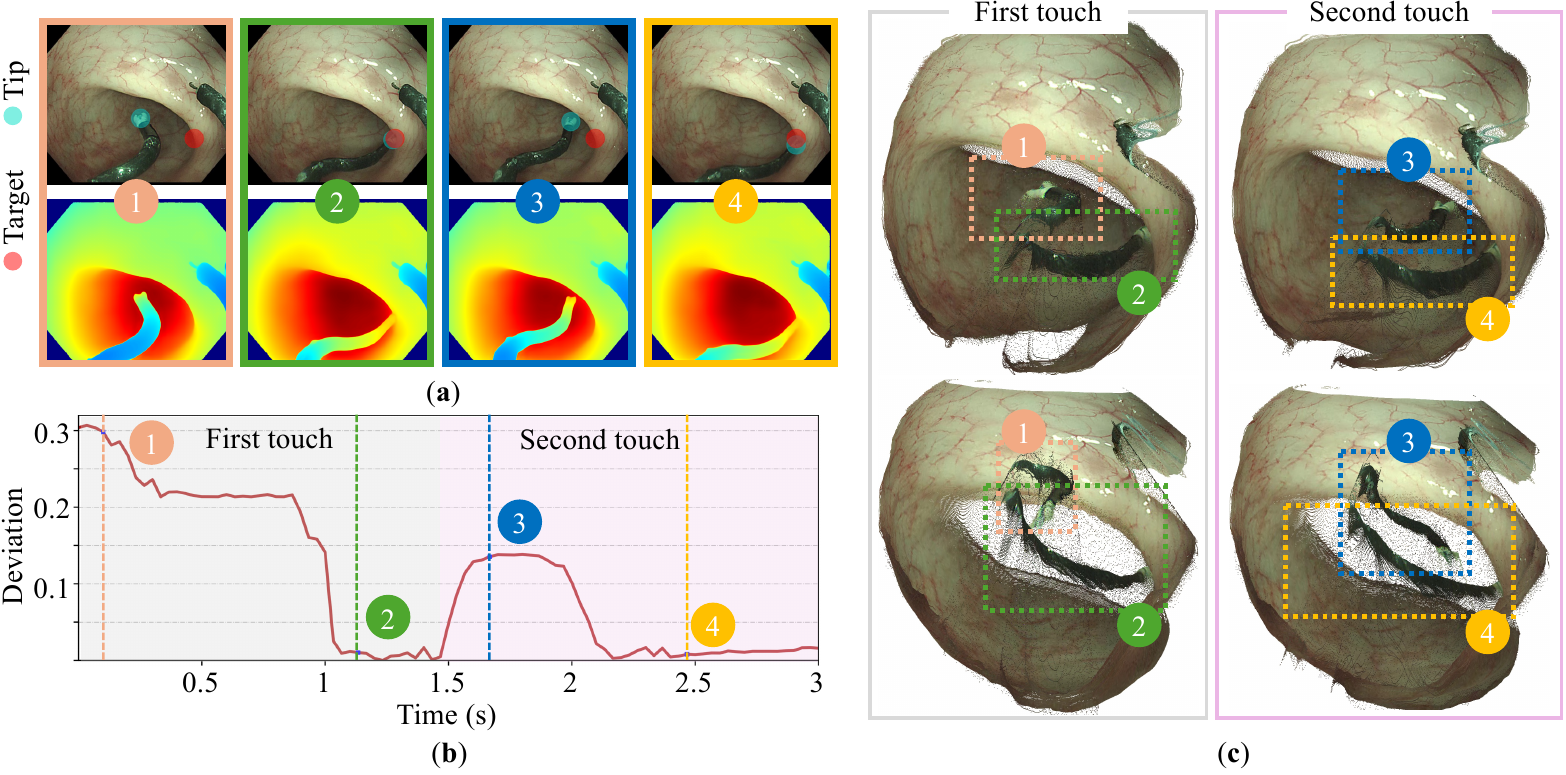"}
    \caption{Analysis of instrument-tissue deviation during flexible instrument touching. (a) Four representative frames along with their corresponding depth maps during the contact events. (b) Variation in distance as the flexible instrument approaches the tissue twice. (c) Multiple views of the 3D structures during the touching events. One demonstration can be found in Video 4.}
    \label{fig:dis_eva}
    \vspace{-0.3cm}
\end{figure*}

\subsection{Instrument-tissue 3D Deviation Inference}
When the surgeon manipulates flexible instruments to contact the colon tissue, it is crucial to establish the distance relationship in 3D space between the instrument tip and the tissue area, as the monocular image can obstruct the surgeon's space analysis. Instrument-tissue deviation inference is achieved through the 3D depth estimation of monocular scenes. This analysis is vital for evaluating the performance of our depth estimation model in surgical scenarios.
As shown in Fig.~\ref{fig:dis_eva}, the surgeon controls the left instrument to touch the tissue twice. During this process, we utilize depth estimation model to compute the distance between the instrument tip and the target tissue. Initially, the surgeon marks both the instrument's tip and tissue area that require deviation analysis. A tracking algorithm is then employed to continuously recognize these marked objects in real-time. Using our depth estimation results, we can calculate the deviation between these two objects. 
Fig.~\ref{fig:dis_eva}(b) presents the variation in distance during the two touching events, showing trends that closely match expected behavior. This indicates that the model effectively captures the dynamics of instrument movement and tissue interaction. Furthermore, the depth maps in Fig.~\ref{fig:dis_eva}(a) illustrate detailed spatial relationships, confirming the model’s ability to detect subtle changes in depth during manipulation. The 3D structures shown in Fig.~\ref{fig:dis_eva}(c) depict how the left instrument’s configuration changes during touching events, enhancing the surgeon’s understanding of instrument positioning relative to the tissue. Overall, these results demonstrate that our depth estimation approach effectively captures instrument-tissue interactions, providing valuable insights for surgical applications and facilitating improved decision-making.

\section{Discussion}
\label{sec:discussion}


This work presents an integrated 2D and 3D perception framework for continuum robotic systems in endoluminal surgery. Our experiments, including module assessments and system-level evaluation, demonstrate significant improvements in the control of flexible instruments and provide comprehensive understanding of complex surgical scenarios. The achieved results validate the effectiveness of our approach in controlled settings, though several limitations and opportunities for advancement warrant further discussion.

\subsection{Limitations}
The primary limitation is the absence of in-vivo human validation; our experiments were conducted on ex-vivo phantoms and synthetic environments. The evaluation tasks—trajectory following and tissue palpation—represent foundational procedures but do not encompass complex manipulations such as tissue resection or biopsy sampling. 
While our perception modules achieve high accuracy, the clinically acceptable error tolerances for different interventions remain unclear. Our monocular depth estimation achieves an error of approximately 4~\textit{mm}, consistent with current state-of-the-art methods in GI scenes~\cite{cui2025learning,wang2024endogslam}. However, different procedures likely have varying accuracy requirements; diagnostic navigation may tolerate larger errors than therapeutic interventions requiring precise tissue manipulation. For reference, structured light-based methods for GI polyp size measurement achieve errors of $\sim$1.5 \textit{mm}~\cite{visentini2018structured}, representing a benchmark for future improvement. Without established clinical benchmarks for adequate depth accuracy in specific GI procedures, it is difficult to assess whether our current performance meets clinical needs or requires further refinement.
Our monocular depth estimation cannot reliably recover geometry in fully occluded regions behind instruments, limiting reasoning about clearance and safe motion planning. This represents a fundamental challenge for vision-based approaches requiring complementary strategies.
Robot kinematics are not currently utilized despite being a valuable information source. Integrating kinematic models with our probabilistic framework through Bayesian fusion would improve robustness when visual features are ambiguous.
Our Unity-based simulator models rigid anatomical structures with realistic textures but does not simulate soft tissue deformation or tool-tissue interaction. While sufficient for validating perception algorithms, this limits evaluation of performance under realistic mechanical interactions—a current field-wide challenge in surgical simulation.

\subsection{Future work}
Critical next steps include in-vivo clinical validation across broader surgical tasks and establishing task-specific accuracy thresholds through expert consultation. Future evaluations should assess functional outcomes (e.g., successful task completion rates) rather than purely technical metrics to better gauge clinical utility.
For depth estimation, several improvements are planned. Explicit illumination modeling with light source position calibration could enable more accurate absolute depth estimation. Alternatively, leveraging known physical dimensions of surgical instruments could provide scale recovery for absolute depth reconstruction. More sophisticated rendering models accounting for complex light-tissue interactions may further reduce estimation errors.
Addressing limitations in occluded region reasoning requires multi-faceted approaches. Temporal fusion and multi-view reconstruction across video sequences could infer geometry in currently occluded regions by integrating past observations as the camera and instruments move. These approaches would enable more robust reasoning about clearance and safe motion planning.
Integrating kinematic models with visual state estimation through physics-informed Bayesian fusion would improve overall system robustness, particularly in scenarios with ambiguous visual features or challenging lighting conditions.
Physics-based simulation with deformable tissue models would generate more realistic training data and enable validation of manipulation tasks involving tissue contact. Integration of the Simulation Open Framework Architecture (SOFA)~\cite{faure2012sofa} into Unity3D via existing interfaces would allow incorporation of finite element modeling for tool-tissue contact simulation. Additionally, reconstructing large-scale 3D organ models from clinical video datasets would enable comprehensive synthetic data generation.
Addressing these limitations will advance the translation of this perception framework into clinical practice.
\section{Conclusion}
\label{sec:conclusion}

In this paper, we proposed an intelligent continuum robotic system for endoluminal surgery, with the goal of enhancing the robustness of robotic procedures through advanced perception algorithms.
We developed three innovative 2D and 3D perception algorithms designed to improve the capabilities of the continuum robotic system. 
These algorithms include a novel segmentation module for accurately identifying the flexible instruments, a probabilistic robot state estimation module that effectively represents instrument's 3D state parameters, and a monocular depth estimation method for measuring distances within the surgical environment.
To support the development of these perception modules, we built a physically-realistic simulator that generates realistic endoluminal surgical scenes and facilitates the collection of substantial data for training.
Extensive experiments were conducted at both the individual module and system levels, demonstrating the effectiveness of the proposed perception algorithms. The precision of flexible instrument segmentation is $>94.5\%$, the robot state estimation achieves a lower angular error ($\sim 6\degree$), and the depth estimation error reaches an RMSE of 4.020\,\textit{mm}. In the robotic system-level evaluation, manipulation time for the trajectory-following task decreased by over 71\% for professionals and around 72\% for inexperienced users due to our perception algorithms. The results indicate significant improvements in the control of the flexible instruments and provide a comprehensive understanding of complex surgical scenarios.
Overall, our work advances the field of robotic endoluminal surgery, paving the way for more precise and reliable surgical interventions.

\bibliographystyle{IEEEtran}
\bibliography{IEEEabrv,ref}



\vspace{-25pt}

\begin{IEEEbiography}[{\includegraphics[width=1in,height=1.25in,clip,keepaspectratio]{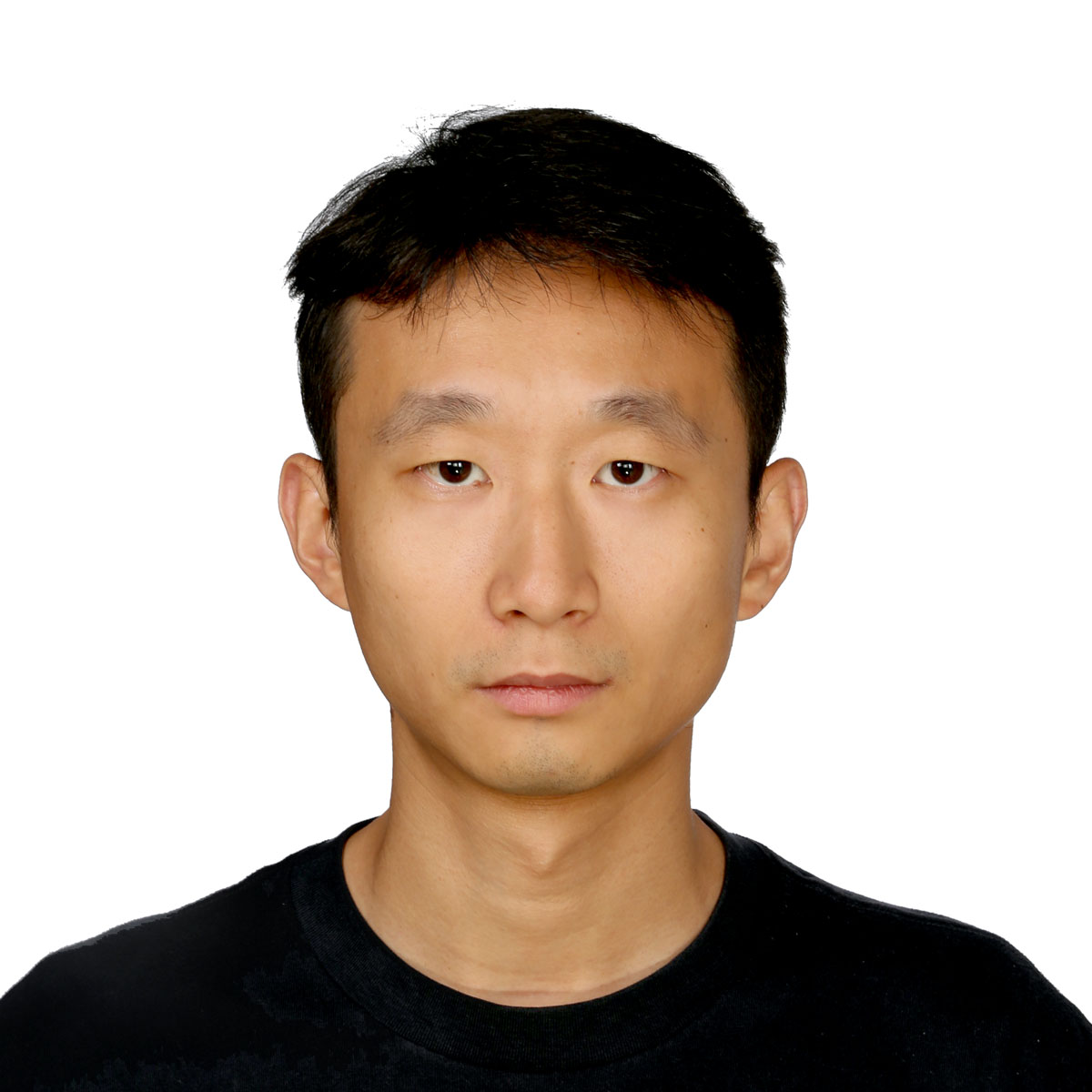}}]{Ruofeng Wei}
received the Ph.D. degree in Biomedical Engineering from City University of Hong Kong, Hong Kong, in 2023. He is currently a Postdoctoral Research Fellow with the Department of Computer Science and Engineering, The Chinese University of Hong Kong, Hong Kong. 
His research interests include surgical robotics, medical image analysis, surgical navigation, automation, and control. 
\end{IEEEbiography}

\vspace{-25pt}

\begin{IEEEbiography}[{\includegraphics[width=1in,height=1.25in,clip,keepaspectratio]{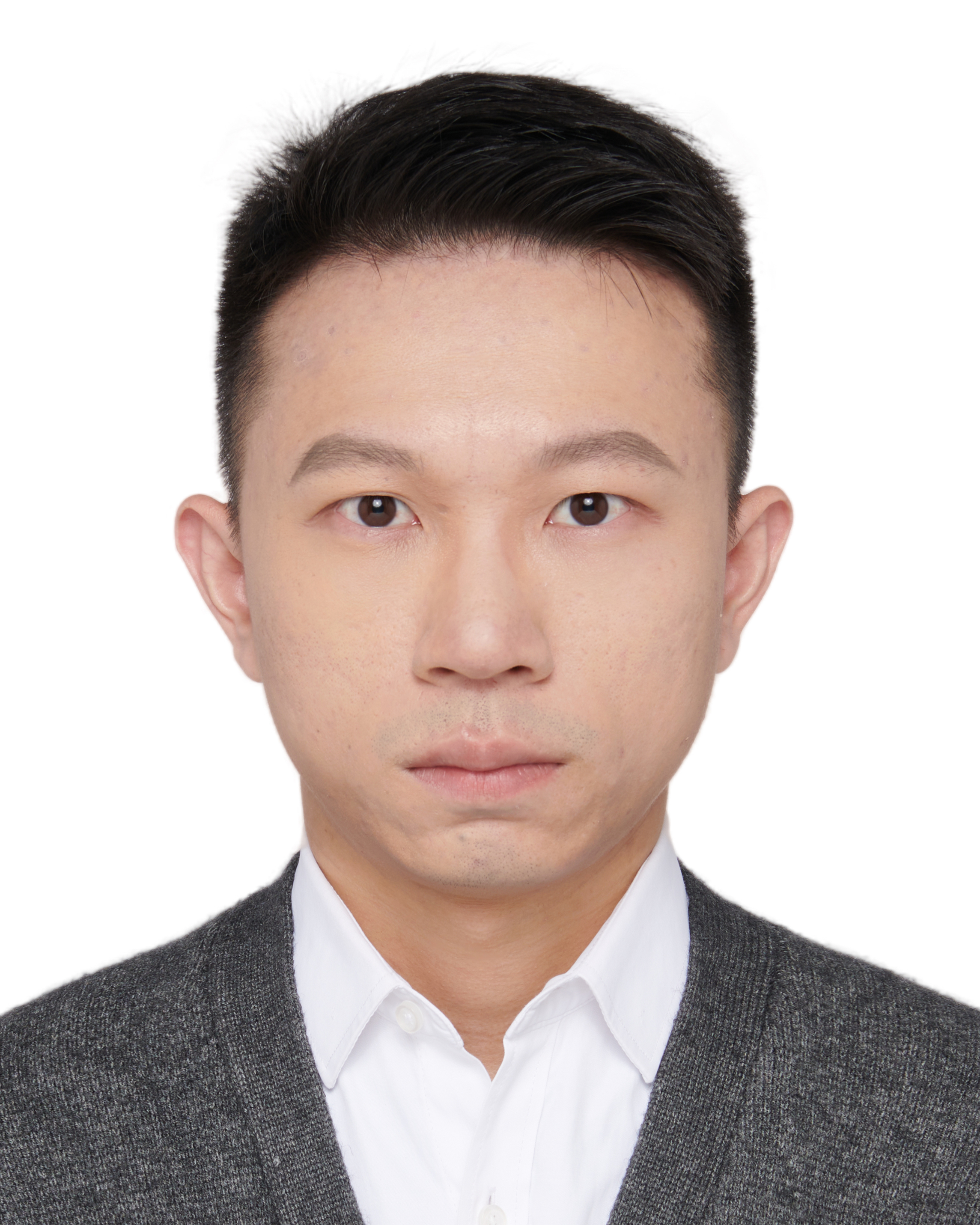}}]{Kai Chen}
received the B.Eng and M.Eng degrees from the School of Remote Sensing and Information Engineering, Wuhan University, Wuhan, China, in 2016 and 2019, and received the Ph.D. degree from the Department of Computer Science and Engineering, The Chinese University of Hong Kong, Hong Kong, in 2024. 
His research interests include 3D computer vision and robot learning.
\end{IEEEbiography}

\vspace{-25pt}

\begin{IEEEbiography}[{\includegraphics[width=1in,height=1.25in,clip,keepaspectratio]{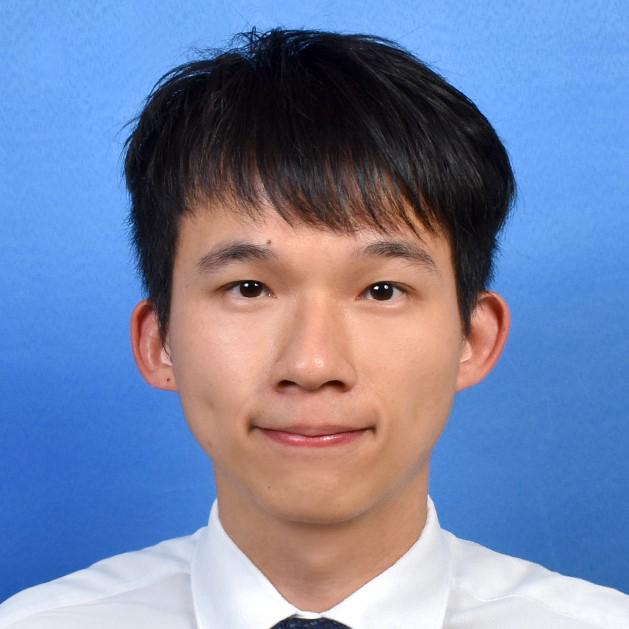}}]{Yui-Lun Ng}
received the B.Sc. degree in mathematics and the M.Sc. degree in big data technology from the Hong Kong University of Science and Technology, Hong Kong, in 2014 and 2018, respectively, and the Ph.D. degree in artificial intelligence and medical image analysis from The University of Hong Kong, Hong Kong, in 2024. 
His research interests include data mining on biomedical data, interpretable machine learning and graph representation learning.
\end{IEEEbiography}

\vspace{-25pt}

\begin{IEEEbiography}[{\includegraphics[width=1in,height=1.25in,clip,keepaspectratio]{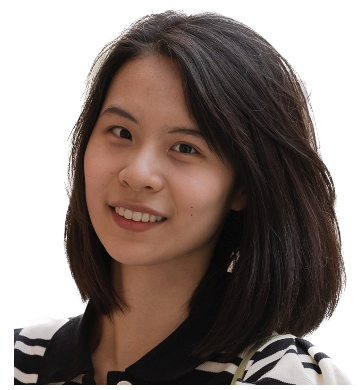}}]{Yiyao Ma}
received the B.E. degree in computer science and engineering from the Beihang University, Beijing, China, in 2023. She is currently working toward the Ph.D. degree in computer science and engineering with the Chinese University of Hong Kong, Hong Kong, China.
Her research interests include flexible robotics, dexterous manipulation, and 3D perceptions. 
\end{IEEEbiography}

\vspace{-25pt}

\begin{IEEEbiography}[{\includegraphics[width=1in,height=1.25in,clip,keepaspectratio]{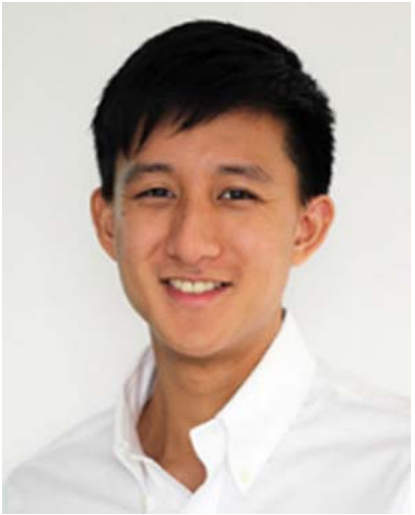}}]{Justin Di-Lang Ho} received the bachelor’s degree in mechatronic engineering from The University of Queensland, Brisbane, QLD, Australia, in 2016, and the M.Phil. degree in the field of robotics from the Department of Mechanical Engineering, The University of Hong Kong, Hong Kong, in 2019. 
His research interests include minimally invasive surgical robotics, including MRI-guided robotics and miniaturized robotic instruments.
\end{IEEEbiography}

\vspace{-20pt}

\begin{IEEEbiography}[{\includegraphics[width=1in,height=1.25in,clip,keepaspectratio]{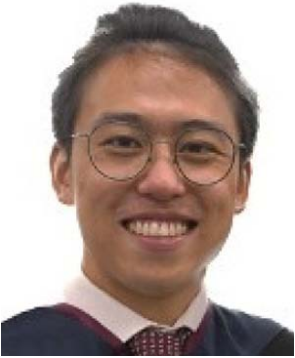}}]{Hon-Sing Tong} received the B.Eng. degree in engineering science and the M.Phil. degree in mechanical engineering from the University of Hong Kong, Hong Kong, in 2019 and 2022, respectively. 
His research interests include surgical navigation, augmented reality-assisted endoscopic surgery, and minimally invasive surgical robotics.
\end{IEEEbiography}

\vspace{-20pt}

\begin{IEEEbiography}[{\includegraphics[width=1in,height=1.25in,clip,keepaspectratio]{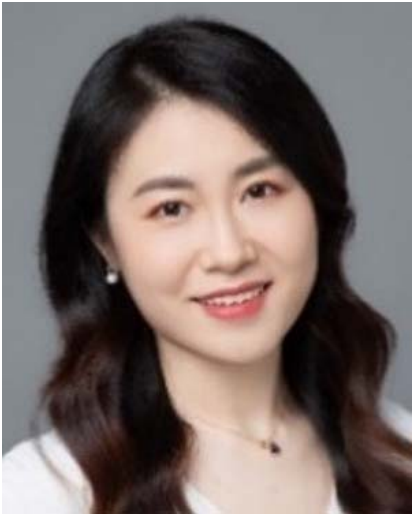}}]{Xiaomei Wang} received the B.E. degree in automation from the Harbin Institute of Technology, Harbin, China, in 2014, the M.E. degree in control science and engineering from the Shenzhen Graduate School, Harbin Institute of Technology, Shenzhen, China, in 2016, and the Ph.D. degree in robotics from The University of Hong Kong, in 2020. 
She was a Postdoctoral Fellow with the Department of Mechanical Engineering, University of Hong Kong, in 2024. 
Her research interests include learning-based robot control and sensing, surgical robotics, and continuum robot control.
\end{IEEEbiography}

\vspace{-20pt}

\begin{IEEEbiography}[{\includegraphics[width=1in,height=1.25in,clip,keepaspectratio]{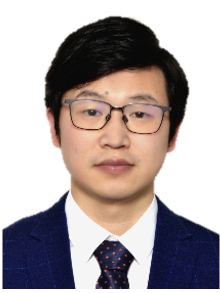}}]{Jing Dai} received the B.Eng. degree from the Wuhan University of Technology, Wuhan, China, in 2017, and the M.S. degree from the Hong Kong University of Science and Technology, Hong Kong, in 2019, and the Ph.D. degree in robotics from the University of Hong Kong, Hong Kong, in 2024. He is currently a Postdoctoral Fellow with the Multi-Scale Medical Robotics Center, InnoHK. His research interests include magnetic resonance imaging (MRI)-guided robotics system.
\end{IEEEbiography}

\vspace{-20pt}

\begin{IEEEbiography}[{\includegraphics[width=1in,height=1.25in,clip,keepaspectratio]{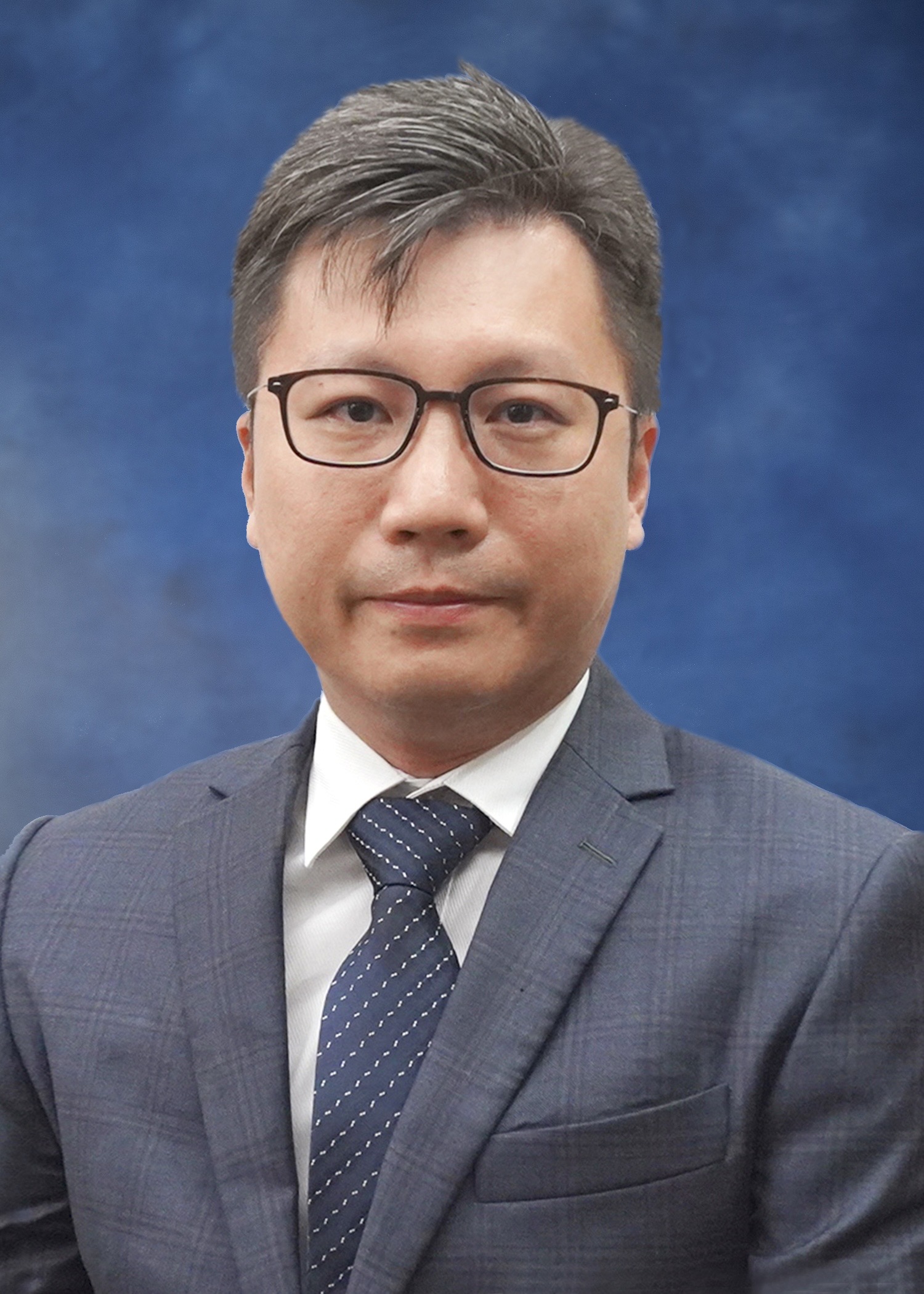}}]
{Ka-Wai Kwok} is a Professor at Mechanical and Automation Engineering, The Chinese University of Hong Kong (CUHK). His research focuses on surgical robotics, intra-operative image processing, and also intelligent and control systems. To date, Ka-Wai has co-authored $>$180 publications with $>$90 clinical fellows and $>$190 engineering scientists. His multidisciplinary work has been recognized by many international publication awards, e.g. 2018 ICRA Best Conference Paper Award and 2020 IROS Toshio Fukuda Young Professional Award (largest flagship academic conferences of robotics). Currently, Ka-Wai is the principal investigator of research group for Interventional Robotic and Imaging Systems (IRIS), which has various inventions licensed/transferred from university to industry in support for their commercialization. He is also a co-founder and director of Agilis Robotics Limited aiming at advancing the interventional endoluminal endoscopy with small, fully flexible robotic instruments and their intelligent control systems. His team successfully performed world’s first transurethral robotic en-bloc resection of bladder tumours (ERBT) in live human in 2024. 
\end{IEEEbiography}

\vspace{-20pt}

\begin{IEEEbiography}[{\includegraphics[width=1in,height=1.25in,clip,keepaspectratio]{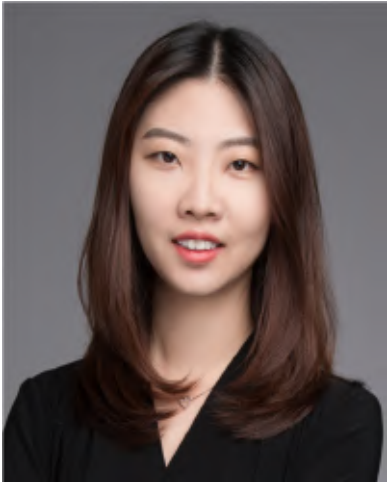}}]{Qi Dou} is currently an Associate Professor at the Department of Computer Science and Engineering at The Chinese University of Hong Kong (CUHK). She is an affiliated member of CUHK T-Stone Robotics Institute, CUHK Institute of Medical Intelligence and XR, Hong Kong Multi-scale Medical Robotics Center, and Hong Kong Centre for Logistics Robotics. Her research is at the interdisciplinary field of AI and robotics technologies for medical applications including autonomous surgical robot, medical imaging, safe embodied AI, etc.
\end{IEEEbiography}


\end{document}